\documentclass{article} 
\usepackage{iclr2026_conference,times}

\usepackage{amsmath,amsfonts,bm}

\def\eqref#1{equation~\ref{#1}}

\def\1{\bm{1}}

\DeclareMathAlphabet{\mathsfit}{\encodingdefault}{\sfdefault}{m}{sl}
\SetMathAlphabet{\mathsfit}{bold}{\encodingdefault}{\sfdefault}{bx}{n}

\usepackage{hyperref}
\usepackage{url}

\usepackage{graphicx}
\usepackage{subcaption}
\usepackage{enumitem}
\usepackage{listings}
\usepackage{booktabs}
\usepackage{multicol}
\usepackage{multirow}
\usepackage{graphicx}
\usepackage{tabularx}
\usepackage{arydshln}
\usepackage{float}
\usepackage{placeins}
\usepackage{xcolor}
\usepackage[hang,flushmargin]{footmisc}
\usepackage[T1]{fontenc}
\usepackage[utf8]{inputenc}
\title{LLM Reasoning for Machine Translation: Synthetic Data Generation over Thinking \mbox{Tokens}}

\author{Armel Zebaze, Rachel Bawden \& Benoît Sagot \\
Inria Paris, France \\
\texttt{armel.zebaze@inria.fr}
}
\iclrfinalcopy 
\begin{document}

\maketitle

\begin{abstract}
Large reasoning models (LRMs) have led to new possibilities in terms of problem-solving,  through the devising of a natural language thought process prior to answering a query. While their capabilities are well known across mathematics and coding tasks, their impact on the task of machine translation (MT) remains underexplored. In this work, we explore the benefits of the generation of intermediate tokens when performing MT across multiple language pairs of different levels of resourcedness and multiple setups. We find that ``thinking tokens'' do not help LRMs better perform MT. This result generalizes to models fine-tuned to reason before translating using distilled chain of thought (CoT) inspired by human translators' practices. Specifically, fine-tuning a model with synthetic CoT explanations detailing how to translate step-by-step does not outperform standard input-output fine-tuning. However, constructing the intermediate tokens by combining the outputs of modular translation-specific prompting strategies results in improvements. Our findings underscore that the contribution of intermediate tokens during fine-tuning highly depends on the presence of translation attempts within them. More broadly, our results suggest that using a teacher to refine target translations or to expand parallel corpora is more impactful than distilling their CoT explanations into ``thinking'' MT models.\footnote{\url{https://github.com/ArmelRandy/llm-reasoning-mt}}
\end{abstract}
%
%

\section{Introduction}
Large Language Models (LLMs) are general-purpose problem solvers \citep{touvron2023llama2openfoundation, openai2024gpt4technicalreport, dubey2024llama3herdmodels, kimiteam2025kimik2openagentic}. Their instruction-following capabilities help them carry out a wide variety of requests provided by users via text. Research on alignment, typically through Reinforcement Learning from Human Feedback (RLHF) \citep{askell2021generallanguageassistantlaboratory, bai2022traininghelpfulharmlessassistant, NEURIPS2022_b1efde53, NEURIPS2023_a85b405e, lambert2025tulu3pushingfrontiers} has greatly contributed to improving the quality of LLMs' outputs. Recently, a new paradigm has emerged: to train LLMs to ``think'' in natural language before answering a query. OpenAI o1 and o3 \citep{o1}, DeepSeek-R1 \citep{deepseekai2025deepseekr1incentivizingreasoningcapability}, Qwen3 \citep{yang2025qwen3technicalreport}, Claude 4 \citep{Claude4} and  Gemini 2.5 \citep{comanici2025gemini25pushingfrontier} \textit{inter alia} are instances of these Reasoning Models (RM) or Thinking Models (TM). They capitalize on RL to generalize the success of Chain-of-Thought (CoT) prompting \citep{NEURIPS2022_9d560961} during training for improved safety robustness and performance. They particularly excel in reasoning-intensive tasks such as olympiad-level mathematics (AIME 2024/2025, HMMT etc.) and competition-level coding \citep{shi2024can, quan2025codeelobenchmarkingcompetitionlevelcode}. When it comes to Machine Translation (MT), they also perform well \citep{chen2025evaluatingo1likellmsunlocking} notably for stylized translation and document-level MT \citep{liu2025newtrendsmodernmachine}.

\begin{figure}[ht]
  \begin{subfigure}{0.1\textwidth}
    \includegraphics[height=5.4cm]{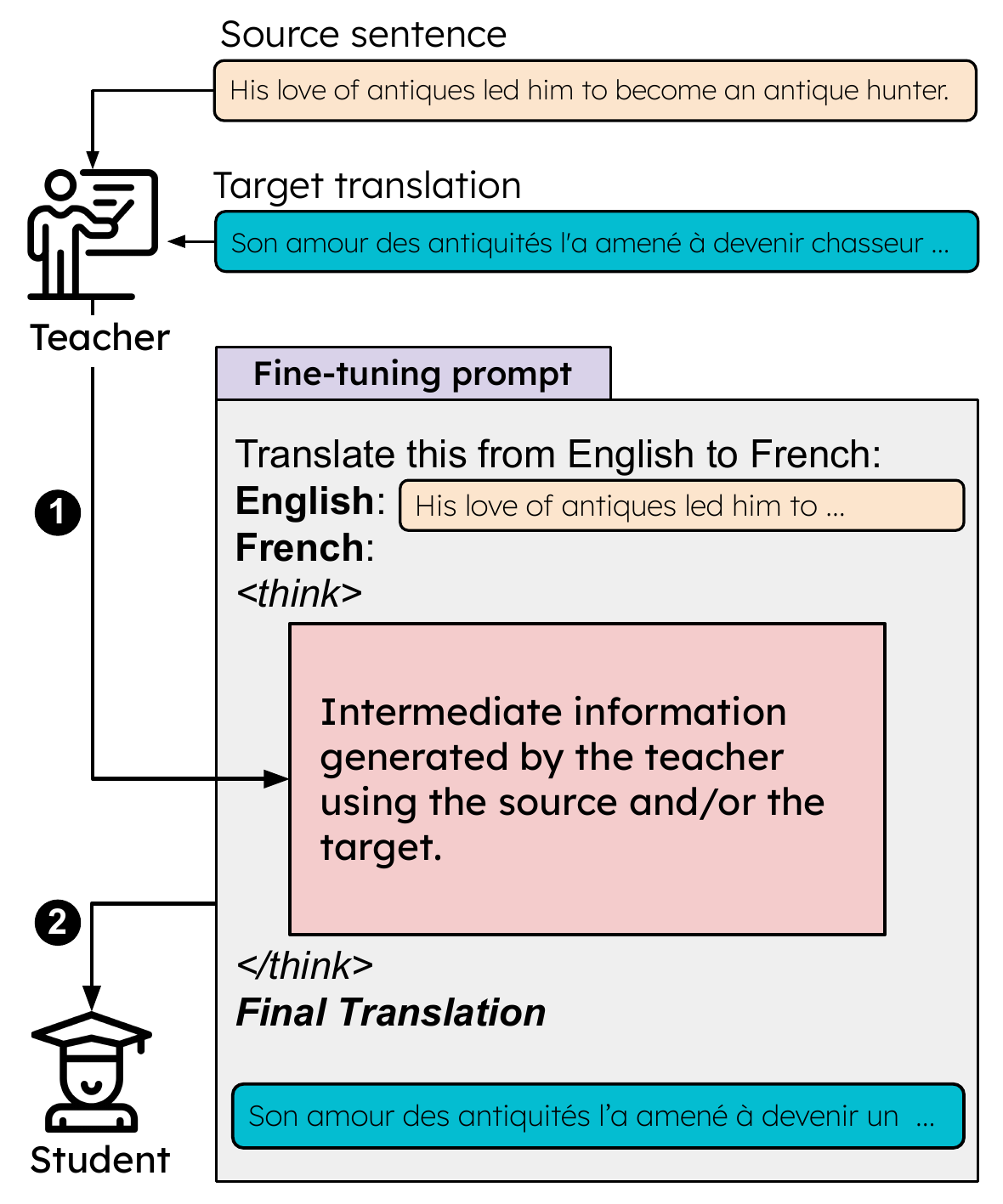}
  \end{subfigure}
  \hskip 3.5cm 
  \begin{subfigure}{0.8\textwidth}
    \includegraphics[height=5.4cm]{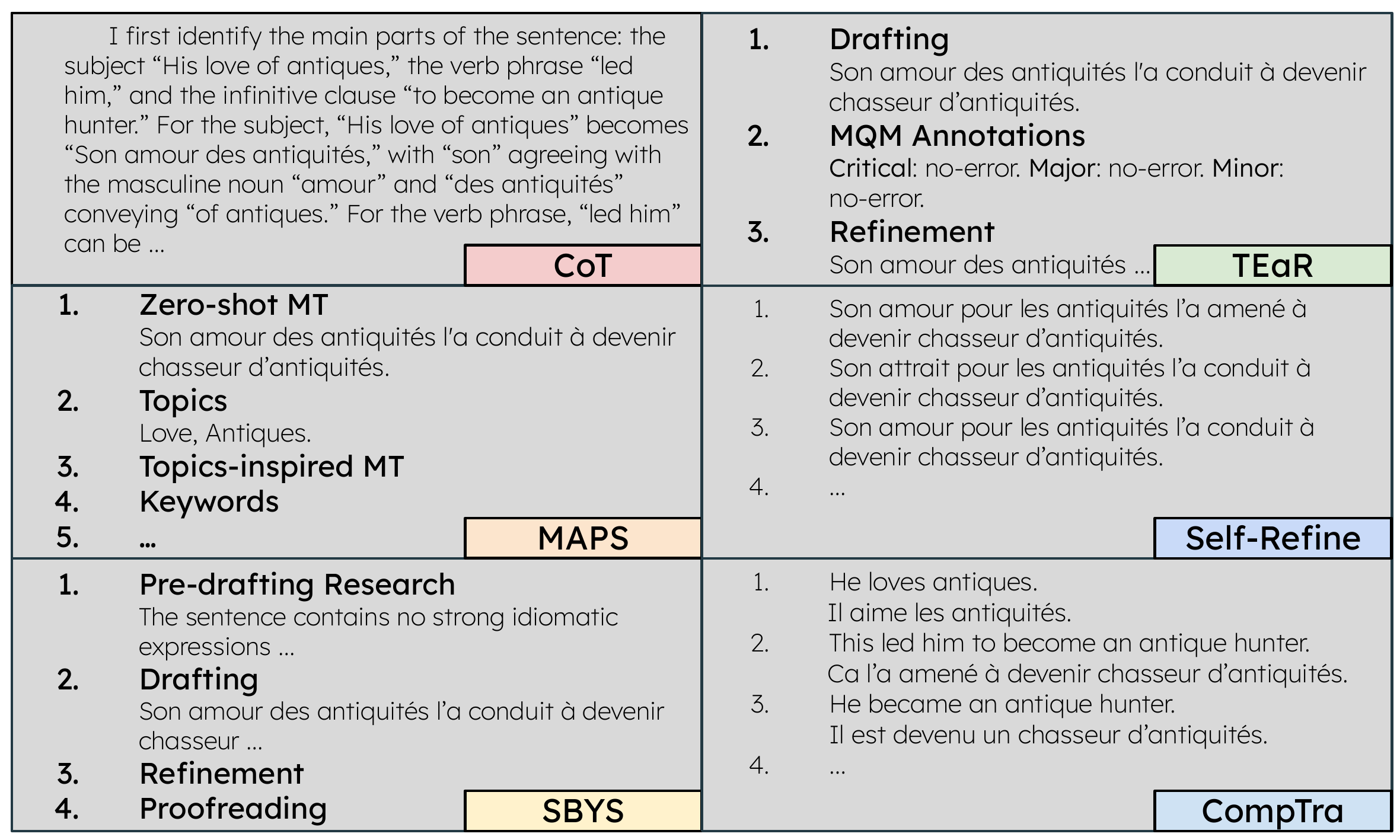}
  \end{subfigure}
  \caption{\textsc{CoT Fine-Tuning} (left): Given a source-target pair, a teacher is prompted to get a thought process on how to obtain the target given the source based on a given strategy (right). The obtained trace is used as intermediate information to fine-tune a student to ``think'' before translating.}
    \label{fig:fig1}
\end{figure}
Labeling o1-like models as \textit{thinking} presupposes that the intermediate tokens they produce before answering meaningfully reflect their reasoning process. Many works challenge this view. For instance,  \citet{bhambri2025interpretabletracesunexpectedoutcomes} find little correlation between the correctness of final answers and the accuracy of intermediate traces, echoing earlier results by \citet{NEURIPS2023_ed3fea90} showing that CoT explanations can be unfaithful. \citet{ma2025reasoningmodelseffectivethinking} further demonstrate that ``not thinking'' can outperform ``explicit reasoning'' on certain challenging tasks, motivating exploration of ``not thinking'' in other settings. In this work, we examine the value of intermediate tokens in MT with LLMs and ask what forms of intermediate information are actually beneficial. MT is a particularly interesting task in this context as CoT prompting has been shown not to result in better translation than vanilla few-shot prompting \citep{nguyen-xu-2025-reasoning}. We show that:
\begin{itemize}
\item{\bf The thinking mode of LRMs does not results in better MT outputs.} We carry out extensive experiments across ten language directions and find no significant benefit from prior thinking. Our experiments cover zero-shot and few-shot settings, three benchmarks, high- and low-resource language pairs (X-to-English and English-to-X directions) and different temperatures of generation, and they all point to the same conclusion.

\item{\bf CoT distillation does not outperform standard fine-tuning.} Many works report a stark improvement of reasoning abilities when fine-tuning a small model to think before answering, using the CoT outputs of a teacher \citep{NEURIPS2022_639a9a17, huang2024o1replicationjourney, li2025llmseasilylearnreason, guha2025openthoughtsdatarecipesreasoning}. We apply this setup to fine-tune LMs to ``think'' before translating and compare it to standard input-output fine-tuning. Our experiments with \texttt{gemma-3-4b-pt} on English to Xhosa suggest the consistent superiority of standard fine-tuning across six different MT-specific CoT templates.

\item{\bf Using traces obtained by translating the source with modular prompting strategies specifically designed for MT outperforms CoT distillation and standard input-output fine-tuning, but ultimately data matters most.} Instead of vanilla CoT distillation, we propose to use the traces obtained when the teacher attempts to translate the source using a modular prompting strategy for MT. Such strategies typically involve an analysis of the source, the proposal of intermediate candidates, and the derivation of the final translation. These steps can be concatenated into a single text, which we then use as intermediate information for fine-tuning the student model. This approach outperforms input-output fine-tuning by up to 3.5 BLEU and 2 MetricX points. Analysis indicates that the gains stem mainly from translation attempts embedded in the traces. We further show that using the teacher to improve the fine-tuning dataset instead, by either enhancing the quality of its target translations or generating additional parallel pairs has greater benefits than relying on thinking tokens, without incurring extra inference cost after fine-tuning.
\end{itemize}

\section{Related Work}\label{sec:related-work}
\paragraph{Reasoning with LLMs.} CoT prompting \citep{NEURIPS2022_9d560961} has revolutionized the approach to reasoning with LLMs. Following In-Context Learning (ICL;\footnote{Also referred to as \textit{few-shot learning}, which is the ability through which LLMs can carry out a wide variety of tasks at inference based on a few demonstrations} \citealp{gpt3}), CoT prompting drives the LLM to explain with natural language the thought process before deriving the solution to a problem. It was shown to be particularly useful for mathematical tasks that require the LLM to think through a set of reasoning steps \citep{cobbe2021trainingverifierssolvemath, hendrycks2021measuring, NEURIPS_DATASETS_AND_BENCHMARKS2021_be83ab3e, suzgun-etal-2023-challenging}. The intuition behind CoT prompting and its success has powered countless related prompting strategies \citep{NEURIPS2022_8bb0d291, zhang2023automatic, yasunaga2024large}. Other developments involve using CoT as a building block to solve sequential problems \citep{zhou2023leasttomost, zebaze-etal-2024-tree}, using CoT in combination with an external tool such as a programming language interpreter \citep{chen2023program, pmlr-v202-gao23f} or to reason on diverse reasoning trajectories \citep{wang2023selfconsistency, NEURIPS2023_271db992, besta2024got, bi2025forestofthought}. CoT-based techniques have also been used to create datasets for supervised fine-tuning \citep{NEURIPS2022_639a9a17, shao2024deepseekmathpushinglimitsmathematical, yue2024mammoth}, which is often subsequent to prior continual pretraining on mathematics and code data \citep{NEURIPS2022_18abbeef, azerbayev2024llemma}. We now have Thinking Models which are trained to ``think'' and produce a long chain of thought before responding to a user query, and these models have set a new state-of-the-art for multiple benchmarks (GPQA \citep{rein2024gpqa}; SWE-Bench \citep{jimenez2024swebench, sweverified}) and keep motivating the creation of more challenging ones \citep{phan2025humanitysexam, chollet2025arcagi2newchallengefrontier}.
Despite their success, \citet{ma2025reasoningmodelseffectivethinking} suggest that these thinking tokens can bring limited gains compared to not thinking\footnote{They induce the ``not thinking mode'' by stopping the thinking process before it even occurs, for example by writing a sentence such as ``Okay, I think I have finished thinking''.} on some reasoning tasks under thinking budget constraints. Others question the informativeness of CoT traces, showing that even incorrect traces can yield correct outcomes \citep{NEURIPS2023_ed3fea90, bhambri2025interpretabletracesunexpectedoutcomes} and that fine-tuning on incorrect traces can be as effective as on correct ones \citep{stechly2025semanticsunreasonableeffectivenessreasonless}. We focus this debate on the specific task of MT and ask whether general-purpose RMs translate better with thinking tokens or if they can be removed entirely.

\paragraph{Machine Translation with LLMs.} MT is one of the many tasks that LLMs can perform via ICL. Historically, encoder-decoder models have been the go-to architecture \citep{NIPS2014_5a18e133, cho-etal-2014-learning, bahdanau2016neuralmachinetranslationjointly, johnson-etal-2017-googles, NIPS2017_3f5ee243}. 
Decoder-based LLMs perform on par with or better than supervised MT models such as NLLB \citep{nllb2022} when dealing with the so-called high-resource languages (HRLs), largely thanks to the availability of large quantities of high quality data on the internet, which facilitates their incorporation in the ever growing pretraining corpora of LLMs. LLMs still struggle with translating from and into low-resource languages (LRLs), but they offer more flexibility when prompting. ICL and the use of few-shot examples (including their selection and order, their number and quality) greatly impact the quality of MT outputs \citep{agrawal-etal-2023-context, moslem-etal-2023-adaptive, hendy2023goodgptmodelsmachine, bawden-yvon-2023-investigating, mu-etal-2023-augmenting, zhu-etal-2024-multilingual, bouthors-etal-2024-retrieving, zebaze-etal-2025-context}, and  various prompting strategies have also been developed for MT such as
``Multi-Aspect Prompting and Selection'' (MAPS; \citealp{he2024exploring}), ``Translating Step-by-Step'' (SBYS; \citealp{briakou-EtAl:2024:WMT}), ``Translate, Estimate, and Refine'' (TEaR; \citealp{feng-etal-2025-tear}), and ``Compositional Translation'' (CompTra; \citealp{zebaze2025compositionaltranslationnovelllmbased}). This includes strategies that iteratively guide LLMs to refine translations, with or without external feedback \citep{chen-etal-2024-iterative, xu-etal-2024-llmrefine, ki-carpuat-2024-guiding}, inspired by the success of similar approaches in reasoning tasks \citep{NEURIPS2023_91edff07, 10.5555/3666122.3666499}. However, standard CoT prompting (e.g.,~\textit{Let's think step by step}) has had little to no success in LLM-based MT, with most works reporting worse results than standard input-output prompting \citep{peng-etal-2023-towards, zebaze2025compositionaltranslationnovelllmbased, nguyen-xu-2025-reasoning}. 
Several works have explored building RMs for MT \citep{wang-etal-2025-drt, wang2025deeptrans, wang2025extrans, he2025r1t1fullyincentivizingtranslation} by closely following what is done for reasoning tasks. They typically prompt a large model (e.g.,~DeepSeek-R1) with a curated CoT prompt that guides it from the source sentence to the target translation, then use the generated CoT for supervised fine-tuning (SFT) followed by RL fine-tuning. However, \citet{zheng2025hunyuanmttechnicalreport} suggests that thinking does do not help MT performance when applying GRPO \citep{shao2024deepseekmathpushinglimitsmathematical} on rewards that only evaluate the final translation. In this work, we focus on SFT and study the pertinence of thinking tokens via CoT distillation, which is already successful on reasoning tasks \citep{huang2024o1replicationjourney, li2025llmseasilylearnreason, guha2025openthoughtsdatarecipesreasoning}. Moreover, given a source, we propose to use modular prompting strategies for MT that have been shown to outperform zero- and/or few-shot MT and ask a teacher to translate the source with the strategies. These prompting strategies resemble a reasoning pipeline, decomposed into multiple steps, each guided by a distinct prompt serving a specific purpose (e.g.~identification of idiomatic expressions, generation of a similar sentence, quality estimation, or drafting). For instance, MAPS \citep{he2024exploring} prompts the LLM to analyze the source and extract three translation-related aspects—keywords, topics, and relevant demonstrations—each used to generate a candidate translation, with the final output selected from these and the zero-shot attempt. SBYS \citep{briakou-EtAl:2024:WMT} begins with a pre-drafting research step, where the LLM identifies idiomatic or otherwise challenging expressions in the source. Based on this analysis, it is then prompted to produce an initial draft translation, followed by a refinement stage. In a subsequent conversation, the LLM is instructed to proofread the refined translation—reflecting on both the source and the previously generated draft. In TEaR \citep{feng-etal-2025-tear}, the LLM first produces a draft translation in a few-shot setting, then generates MQM-style error annotations and refines the draft accordingly. Self-Refine \citep{chen-etal-2024-iterative} involves drafting an initial translation and iteratively improving it through self-feedback. In CompTra \citep{zebaze2025compositionaltranslationnovelllmbased}, the LLM decomposes the source into smaller phrases, translates them independently in a few-shot manner, and uses these synthetic pairs as additional demonstrations to improve the final translation. The modularity of these methods lies in their multi-step structure. The outputs (or traces) of these individual steps can be concatenated into a single text and used as a CoT (intermediate tokens) to fine-tune a student, thereby building a thinking MT model, i.e.~a model that thinks before translating.


\section{Benchmarking LRMs at Scale: To Think, or Not To Think?}

We first investigate the influence of prior reasoning on the translation quality in general-purpose thinking models. We compare two conditions: (i)~the model is allowed to generate reasoning tokens prior to producing the translation (up to 3500 tokens), and (ii)~reasoning is explicitly suppressed by appending \texttt{<think>\textbackslash n\textbackslash n</think>} to the prompt (Non-Thinking Mode; \citealp{yang2025qwen3technicalreport}). We carry out experiments with the Qwen3 model family \citep{yang2025qwen3technicalreport}, ranging in size from 0.6B to 32B parameters, in a zero-shot English-to-X setting for ten FLORES-200 languages: Czech, Finnish, French, German, Japanese, Kazakh, Lithuanian, Portuguese, Spanish, and Turkish. The results, summarized in Table~\ref{tab:zs_hrl} show that the performance with and without prior thinking is similar. Non-thinking is slightly better, in particular in terms of MetricX but the difference is usually less than 0.5 MetricX point. We provide additional results with more models and directions in Appendix~\ref{appendix:additional_experiments}, and with two other benchmarks, NTREX~128 \citep{federmann-etal-2022-ntrex, barrault-etal-2019-findings} and TICO-19 \citep{anastasopoulos-etal-2020-tico}, in Appendix~\ref{appendix:ntrex_and_tico}. 
%
\begin{table*}[ht]
\vskip 0.15in
\small
\begin{center}
\resizebox{\textwidth}{!}{
\begin{tabular}{lrrrrrrrrrrrrrr}
\toprule
\multirow{2}{*}{Models}  & \multicolumn{2}{c}{Czech} & & \multicolumn{2}{c}{Finnish} & & \multicolumn{2}{c}{French} & & \multicolumn{2}{c}{German} & & \multicolumn{2}{c}{Japanese}\\
\cmidrule{2-3} \cmidrule{5-6} \cmidrule{8-9} \cmidrule{11-12} \cmidrule{14-15}
{} & {BLEU} & {MetricX} &  & {BLEU} & {MetricX} & & {BLEU} & {MetricX} & & {BLEU} & {MetricX} & & {BLEU} & {MetricX}\\
\midrule
\textsc{Qwen3-0.6B}           & \bf 4.63 & 22.86 & & 2.23 & 23.64 & & \bf 24.65 & 9.03 & & \bf 16.00 & 9.95 & & \bf 7.65 & 8.67 \\
\textit{\quad w/o Thinking}   & 3.99 & \bf 13.60 & & \bf 2.80 & \bf 21.16 & & 23.33 & \bf 8.52 & & 15.27 & \bf 8.21 & & 5.71 & \bf 7.56 \\
\midrule
\textsc{Qwen3-1.7B}           & 14.73 & 15.58 & & 5.83 & 20.29 & & 37.98 & 4.80 & & 27.86 & 4.50 & & \bf 15.49 & 5.84 \\
\textit{\quad w/o Thinking}   & \bf 15.51 & \bf 15.09 & & \bf 7.08 & \bf 19.38 & & \bf 38.08 & \bf 4.44 & & \bf 28.03 & \bf 3.97 & & 15.08 & \bf 5.87  \\
\midrule
\textsc{Qwen3-4B}             & 23.82 & \bf 9.05 & & 13.47 & \bf 14.22 & & \bf 45.40 & 3.19 & & \bf 35.06 & 2.58 & & 19.87 & 5.06  \\
\textit{\quad w/o Thinking}   & \bf 24.43 & 9.27 & & \bf 13.59 & 14.30 & & 44.69 & \bf 3.14 & & 34.64 & \bf 2.48 & & \bf 20.42 & \bf 4.82 \\
\midrule
\textsc{Qwen3-8B}             & 30.11 & \bf 6.38 & & \bf 19.29 & \bf 9.95 & & 48.72 & 2.59 & & \bf 39.29 & 1.62 & & 23.45 & 4.31  \\
\textit{\quad w/o Thinking}   & \bf 30.27 & 6.61 & & 19.21 & 10.20 & & \bf 49.03 & \bf 2.49 & & 39.05 & \bf 1.56 & & \bf 24.43 & \bf 4.08 \\
\midrule
\textsc{Qwen3-14B}            & \bf 34.07 & \bf 4.99 & & 22.73 & 7.74 & & 51.21 & 2.26 & & \bf 42.39 & 1.29 & & 26.25 & 3.80  \\
\textit{\quad w/o Thinking}   & 33.55 & 5.16 & & \bf 23.17 & \bf 7.67 & & \bf 51.88 & \bf 2.12 & & 41.64 & \bf 1.16 & & \bf 27.31 & \bf 3.75 \\
\midrule
\textsc{Qwen3-32B}            & \bf 34.62 & 4.70 & & \bf 24.33 & 7.25 & & \bf 51.80 & \bf 2.01 & & \bf 42.78 & \bf 1.09 & & 26.44 & 3.88  \\
\textit{\quad w/o Thinking}   & 33.27 & \bf 4.64 & & 24.14 & \bf 7.05 & & 50.94 & \bf 2.00 & & 42.08 & \bf 1.07 & & \bf 27.33 & \bf 3.62  \\
\end{tabular}
}
\resizebox{\textwidth}{!}{
\begin{tabular}{lrrrrrrrrrrrrrr}
\toprule
\multirow{2}{*}{Models}  & \multicolumn{2}{c}{Kazakh} & & \multicolumn{2}{c}{Lithuanian} & & \multicolumn{2}{c}{Portuguese} & & \multicolumn{2}{c}{Spanish} & & \multicolumn{2}{c}{Turkish}\\
\cmidrule{2-3} \cmidrule{5-6} \cmidrule{8-9} \cmidrule{11-12} \cmidrule{14-15}
{} & {BLEU} & {MetricX} &  & {BLEU} & {MetricX} & & {BLEU} & {MetricX} & & {BLEU} & {MetricX} & & {BLEU} & {MetricX}\\
\midrule
\textsc{Qwen3-0.6B}           & 0.41 & 23.61 & & 1.28 & 24.50 & & \bf 25.58 & 9.11 & & \bf 17.80 & 7.67 & & 5.88 & 21.66 \\
\textit{\quad w/o Thinking}   & \bf 0.49 & \bf 22.26 & & \bf 1.77 & \bf 23.78 & & \bf 23.31 & \bf 8.68 & & 17.00 & \bf 7.04 & & \bf 6.27 & \bf 20.00 \\
\midrule
\textsc{Qwen3-1.7B}           & 0.92 & \bf 23.41 & & 5.03 & \bf 21.70 & & \bf 39.02 & 4.46 & & \bf 25.43 & 3.87 & & 13.20 & 15.48  \\
\textit{\quad w/o Thinking}   & \bf 1.36 & 23.54 & & \bf 5.83 & 21.56 & & \bf 39.01 & \bf 4.35 & & \bf 25.45 & \bf 3.75 & & \bf 14.35 & \bf 13.94 \\
\midrule
\textsc{Qwen3-4B}             & 8.02 & \bf 16.02 & & 12.83 & 15.69 & & \bf 46.08 & \bf 2.97 & & 28.46 & \bf 2.69 & & 21.49 & 9.75 \\
\textit{\quad w/o Thinking}   & \bf 8.15 & 16.37 & & \bf 13.05 & \bf 15.45 & & 45.80 & 3.09 & & \bf 28.70 & \bf 2.69 & & \bf 21.74 & \bf 9.55 \\
\midrule
\textsc{Qwen3-8B}             & \bf 13.76 & \bf 11.68 & & 17.49 & 11.51 & & \bf 48.93 & 2.53 & & \bf 30.77 & 2.15 & & \bf 27.09 & 7.28 \\
\textit{\quad w/o Thinking}   & 12.98 & 11.76 & & \bf 18.03 & \bf 11.30 & & 48.76 & \bf 2.41 & & 30.60 & \bf 2.03 & & 26.90 & \bf 6.76  \\
\midrule
\textsc{Qwen3-14B}            & \bf 17.81 & \bf 8.87 & & 22.06 & 8.78 & & 50.47 & 2.25 & & 31.77 & 1.95 & & \bf 30.72 & 5.93 \\
\textit{\quad w/o Thinking}   & 17.42 & \bf 8.84 & & \bf 22.82 & \bf 8.45 & & \bf 51.42 & \bf 2.09 & & \bf 32.02 & \bf 1.86 & & 29.85 & \bf 5.70 \\
\midrule
\textsc{Qwen3-32B}            & 17.33 & 9.54 & & 23.70 & 8.41 & & 51.01 & \bf 1.96 & & 31.84 & \bf 1.66 & & \bf 31.68 & 5.80 \\
\textit{\quad w/o Thinking}   & \bf 18.13 & \bf 8.39 & & \bf 24.10 & \bf 7.58 & & \bf 51.59 & \bf 1.96 & & \bf 32.18 & \bf 1.71 & & 30.48 & \bf 5.59 \\
\bottomrule
\end{tabular}
}
\end{center}
\caption{BLEU and MetricX scores for ten English $\rightarrow$ X directions from FLORES~200\iffalse~\citep{goyal-etal-2022-flores, nllb2022}\fi, with thinking (first line) and without thinking (second line). Best results are highlighted in bold.}
\label{tab:zs_hrl}
\end{table*}
\begin{figure}[ht]
\begin{center}
    \includegraphics[width=\linewidth]{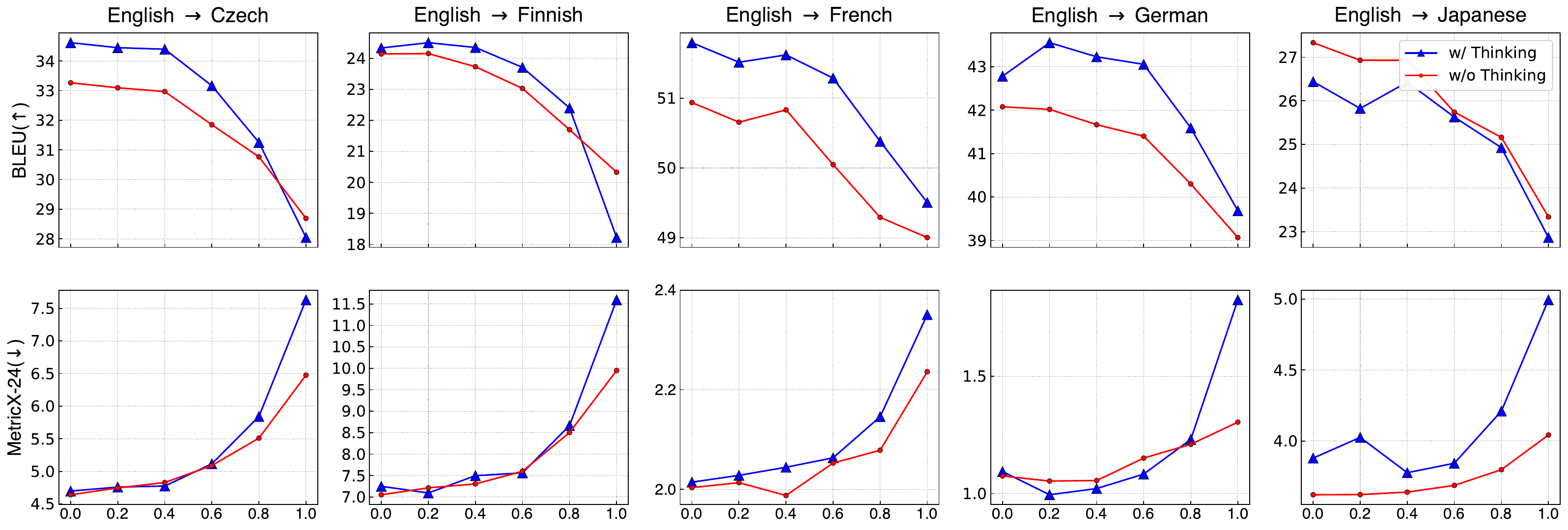}
    \caption{Impact of the Temperature on the translation quality with and without thinking tokens.}
    \label{fig:temperature}
\end{center}
\end{figure}

We evaluate \texttt{Qwen3-32B} when sampling with temperatures 0.2, 0.4, 0.6, 0.8 and 1.0 and plot the results in  Figure~\ref{fig:temperature}. As opposed to default usage recommendations for TMs, sampling degrades the performance, confirming the results obtained by \citet{chen2025evaluatingo1likellmsunlocking} with \texttt{DeepSeek-R1-670B}. In addition, at each temperature, outputting thinking tokens or not gives approximately the same level of performance. We observe a slight performance gap in favor of thinking tokens with respect to BLEU ($\leq$ 1 BLEU point), but they are behind with respect to MetricX. Low temperatures correlate with higher performance and in particular $T=0.0$ works best in general. The results of these experiments suggest that MT does not significantly benefit from the presence of thinking tokens. 

\section{Approaches to Improving MT with Intermediate Reasoning}

%
Given that general-purpose RMs do not seem to benefit from outputting thinking tokens prior to translation, we investigate how to build a successful ``thinking'' MT model, i.e.,~one that first produces intermediate reasoning before translation and outperforms models trained without intermediate steps. 
To this end, we apply \textsc{CoT Fine-Tuning (CoTFT)}, in which a student model is trained to first produce intermediate tokens (that we call CoT or ``thoughts'') before generating the final target translation, as shown in Figure~\ref{fig:fig1}. Using a parallel dataset and a teacher model, we explore what types of intermediate information can be generated by the teacher to train a ``thinking'' MT model (student). There are multiple ways of generating intermediate information with a teacher (see the right side of Figure~\ref{fig:fig1}), which we categorize into two types.
\begin{itemize}
    \item \textbf{CoT prompting.} This corresponds to the standard CoT distillation approach inherited from reasoning tasks (Figure~\ref{fig:fig1}, right, first box). For each source–target pair, the teacher is fed with a curated CoT prompt inspired by human translation strategies. It produces a reasoning trace explaining how to obtain the target from the source, or justifying why the given target is a correct translation of the source. In doing so, the model emulates the strategies used by human translators. It produces a first-person thought process in which it explains how it analyzes the sentence—identifying elements such as the subject, verb, and object—and how it arrives at the target translation by reasoning about linguistic aspects (syntactic rules, word order) and the broader context. Further details are provided in Appendix~\ref{appendix:cot_construction}.
    \item \textbf{Modular translation-specific prompting strategies.} Instead of adopting the classical road, we propose using as intermediate information the traces obtained after applying modular translation-specific prompting strategies to translate the source. As mentioned in Section~\ref{sec:related-work}, they generally involve multiple steps, (see the five other boxes of Figure~\ref{fig:fig1}):
    \begin{itemize}
        \item \textbf{MAPS}: a modular process comprising source analysis (extraction of keywords, topics, and relevant demonstrations) and corresponding translation attempts (each inspired by the extracted information), complemented by zero-shot translation.
        \item \textbf{SBYS}: a four-step process comprising pre-drafting research (identification of expressions that may pose a challenge for translation), drafting, refinement, and proofreading (for terminology, fluency, etc.).
        \item \textbf{TEaR}: a three-step process comprising translation (in a few-shot setting), annotation (of potential translation errors), and refinement (based on these annotations).
        \item \textbf{Self-Refine}: an iterative process comprising the initial translation (in a zero-shot setting) and successive rounds of self-refinement (to improve accuracy and fluency).
        \item \textbf{CompTra}: a three-step process comprising decomposition of the source into simpler phrases, translation of each phrase (in a few-shot setting), and recombination\footnote{We do not use the output of the recombination step when building the intermediate tokens.} (into a final translation).
    \end{itemize}
     Given a source sentence, we apply all the steps of each selected prompting strategy for MT and concatenate the outputs to form a text, which serves as intermediate information for \textsc{CoTFT}. We aim to determine whether this approach results in improved MT models, analogous to how RMs generalize the superiority of CoT prompting over direct input–output prompting with standard LLMs.
\end{itemize}
We compare \textsc{CoTFT} to \textsc{Input-Output Fine-Tuning (IOFT)}, the baseline approach where the student is trained to directly predict the target translation given a source sentence. In both cases, the source and target are the same and the difference only lies in the presence of intermediate reasoning during training.
In summary, we first examine whether prompting a teacher to emulate a human translator and produce CoT traces helps to produce thinking MT model (student) that are more effective than a standard IOFT model.

\section{Experiments}

\subsection{Experimental Setup} \label{section:experimental_setup}

\paragraph{Evaluation Datasets.} Our main evaluation dataset is FLORES-200~\citep{goyal-etal-2022-flores, nllb2022} devtest set (1012 examples). 
For fine-tuning (distillation experiments), we focus on two languages: Xhosa, an LRL, in the main paper, and Lithuanian, a HRL, in the appendix.

\paragraph{Fine-tuning Datasets.} For Xhosa, we use \texttt{Llama-4-Scout-17B-16E-Instruct} \citep{MetaAI_LLaMA-4} and synthetic, multi-domain sentence-level data generated using the \textsc{TopXGen} pipeline \citep{zebaze2025topxgentopicdiverseparalleldata}.\footnote{The pipeline enables the generation of English–LRL parallel data (in this cases the LRL being Xhosa) from LLM-generated LRL texts, backtranslated into English as a way of alleviating the scarcity of diverse, high-quality datasets for LRLs).}$^,$
\footnote{\url{https://hf.co/datasets/almanach/topxgen-llama-4-scout-and-llama-4-scout}} 
For Lithuanian, we use the WMT19 dataset \citep{barrault-etal-2019-findings} for training and run the same experiments as for Xhosa, as detailed in Appendices~\ref{appendix:cot_distillation_lt}{} to~\ref{appendix:sentence_decomposition_lt}.

\paragraph{Models.} For Xhosa, we use \texttt{Llama-4-Scout-17B-16E-Instruct} \citep{MetaAI_LLaMA-4} as the teacher (to generate reasoning traces for \textsc{CoTFT}) and \texttt{gemma-3-4b-pt} as the student. Ablation studies additionally consider \texttt{gemma-3-27b-it} \citep{gemmateam2025gemma3technicalreport} and \texttt{DeepSeek-R1-Distill-Llama-70B} \citep{deepseekai2025deepseekr1incentivizingreasoningcapability} as alternative teachers. For Lithuanian, we pair \texttt{gemma-3-27b-it} as the teacher with \texttt{gemma-3-1b-pt} as the student.

\paragraph{Evaluation Metrics.}
Our main evaluation metric is MetricX-24~\citep{juraska-etal-2024-metricx}. We use the reference-based version \texttt{MetricX-24-Hybrid-XXL}, which supports the same 101 languages as mT5~\citep{xue2021mt5}. MetricX assigns a score ranging from 0 to 25, with higher scores indicating more errors in the translation. We also evaluate using BLEU\footnote{nrefs:1$|$case:mixed$|$eff:no$|$tok:flores200$|$smooth:exp$|$version:2.4.2}~\citep{10.3115/1073083.1073135} as implemented in sacreBLEU~\citep{post-2018-call}.

\paragraph{Implementation Details}
We fine-tune all our models for 5k steps on one H100 80G with a learning rate of 1e-5, a constant scheduler with 500 warm-up steps (from 1e-6) and a batch size of 4. For \textsc{IOFT} we use 4 gradient accumulation steps and a maximum sequence length equal to 512, whereas for \textsc{CoTFT} we use 16 gradient accumulation steps and a maximum sequence length of 2048. All models are evaluated in a zero-shot fashion with greedy decoding unless stated otherwise. See Appendix~\ref{appendix:resources} and \ref{appendix:details} for additional details.

\subsection{Distilled Chain-of-Thought as intermediate tokens} \label{subsection:cot_distillation}

\begin{figure}[ht]
\begin{center}
    \includegraphics[width=\linewidth]{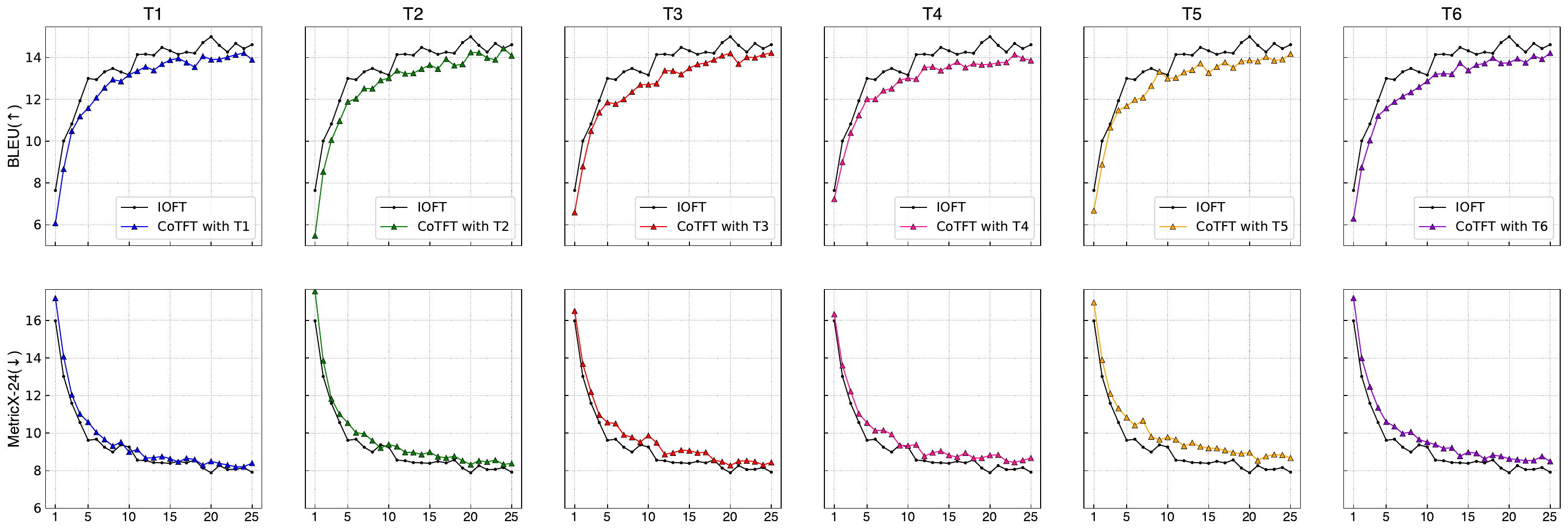}
    \caption{Comparison between \textsc{IOFT} and \textsc{CoTFT} with six different CoT templates. Across all figures, each unit on the x-axis represents 200 steps.}
    \label{fig:coft}
\end{center}
\end{figure}

We compare \textsc{IOFT} and \textsc{CoTFT} in the CoT distillation setup when the teacher is \texttt{Llama-4-Scout-17B-16E-Instruct}. We evaluate each of the six ``CoT instance construction'' prompt templates reflecting human translators' reasoning proposed by \citet{feng2025mtr1zeroadvancingllmbasedmachine} for generating cold-start data for their \texttt{R1-T1} model: Hierarchical translation, Triangulating translation, Backtranslation, Context-aware translation, translation explanation and structural transformation (see Appendix~\ref{appendix:cot_construction}). We fine-tune \texttt{gemma-3-4b-pt} and compare the performance of all six \textsc{CoTFT} variants against \textsc{IOFT}. It is important to recall that in all scenarios, the source and target are the same, only presence or absence of traces and their template change. We report the BLEU and MetricX scores on FLORES-200 every 200 steps in Figure~\ref{fig:coft}. We observe that \textsc{CoTFT} consistently fails to improve over \textsc{IOFT} (in black) across all templates. The variability of performance across templates is negligible; they fall short compared to \textsc{IOFT} by about 0.5 BLEU and 0.5 MetricX points. We ran the same experiment with \texttt{DeepSeek-R1-Distill-Llama-70B} as the teacher and reached the same conclusions (see Appendix~\ref{appendix:thinking_deepseek}).


\subsection{MT traces generated by prompting strategies as intermediate tokens}
RMs were built based on the premise that a ``thinking process'' formalized with natural language could help achieve better results. Asking an LLM to translate a sentence step-by-step does not improve over CoT-free zero-shot MT. However multistep prompting strategies that mimic translation reasoning exist. Given a teacher and a prompting strategy, can the traces generated during translation, when used as intermediate information, help produce better outputs? 
We consider five modular prompting strategies: MAPS, SBYS, TEaR, Self-Refine and CompTra.
%
%
%
\begin{figure}[t]
\begin{center}
    \includegraphics[width=\linewidth]{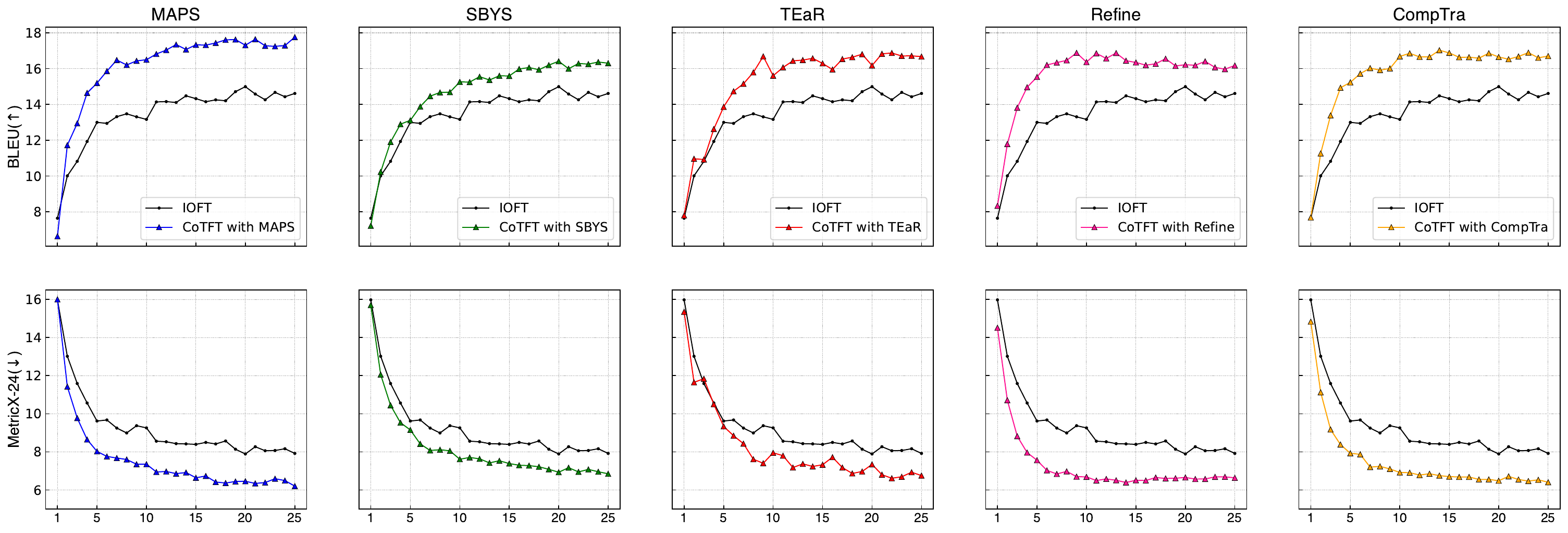}
    \caption{Comparison between \textsc{IOFT} and \textsc{CoTFT} with five different prompting strategies.}
    \label{fig:prompting}
\end{center}
\end{figure}
As shown in Figure~\ref{fig:prompting}, \textsc{CoTFT} on traces based on MT prompting strategies outperforms \textsc{IOFT}. For instance, we reach up to 3.5 BLEU and 2.0 MetricX gains with MAPS. For the other prompting strategies, improvements remain around +2 BLEU and -1.5 MetricX. Using CoT traces derived from these strategies appears beneficial—but why? The key difference is that, unlike pure CoT prompting, most of these strategies (except CompTra) include one or multiple drafting phases. The success of \textsc{CoTFT} may therefore stem from drafts that surpass the ground truth. 
We test this hypothesis as follows.
%

For each strategy, we use the quality estimation score BLASER~2.0-QE \citep{duquenne2023sonarsentencelevelmultimodallanguageagnostic, dale-costa-jussa-2024-blaser} to obtain the best translation between the ground truth and the attempts embedded in the teacher's traces.
We consider 2 scenarios. \textsc{IOFT-Max(strategy)} which is \textsc{IOFT} where the target is replaced by the best one between the ground truth  and those potentially generated by the prompting strategy. \textsc{CoTFT-Max(strategy)} which is analogous to \textsc{IOFT-Max(strategy)} but with the intermediate tokens.
In addition to the above scenarios, we consider \textsc{IOFT-BoA} (best of all) which is \textsc{IOFT} where the target is the best between the ground truth and translations embedded into the traces obtained across all the prompting strategies considered (MAPS, SBYS, TEaR, Self-Refine and CompTra).

\begin{figure}[ht]
\begin{center}
    \includegraphics[width=\linewidth]{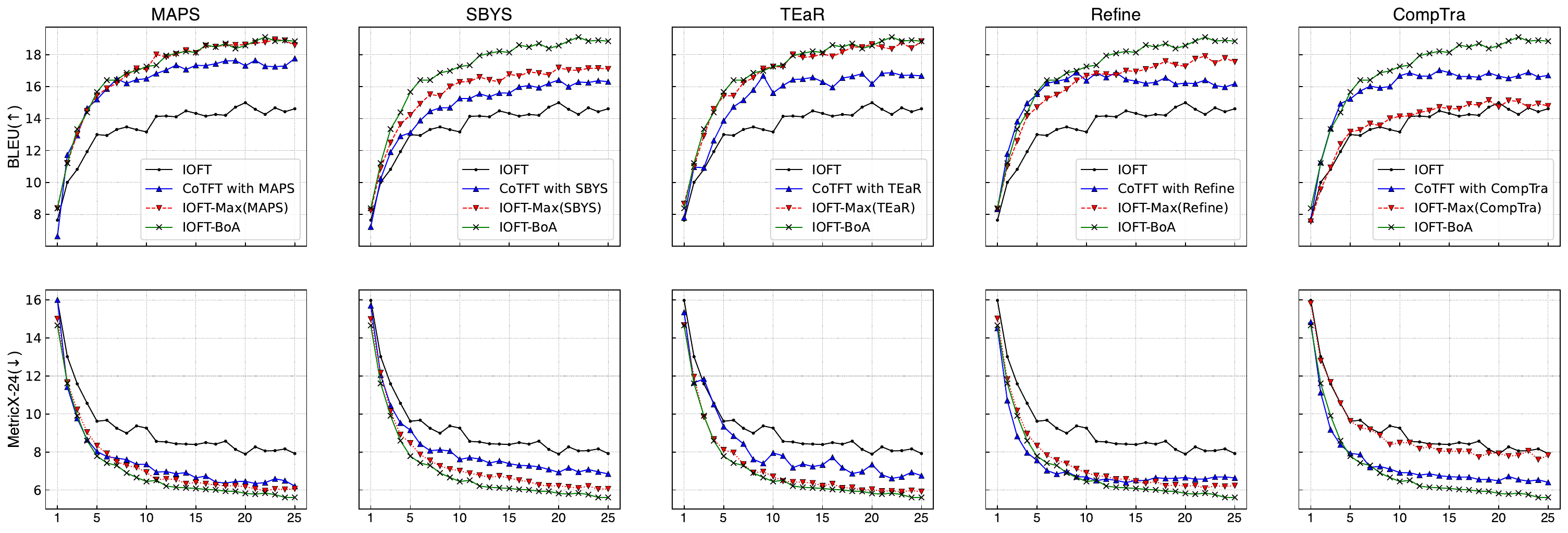}
    \caption{Comparison between \textsc{IOFT} and \textsc{CoTFT} with six different prompting strategies.}
    \label{fig:prompting-v2}
\end{center}
\end{figure}

\paragraph{First scenario.} For MAPS, SBYS, TEaR and Self-Refine, \textsc{IOFT-Max} (in red) works better than \textsc{CoTFT} (i.e. with the traces) and \textsc{IOFT} (Figure~\ref{fig:prompting-v2}). This indicates that the quality of the target is an important factor for downstream performance. Using better ground truths (\textsc{IOFT-BoA}, in green) can make \textsc{IOFT} go from 14 BLEU to 18 BLEU (8 MetricX to 5.6 MetricX) with the same number of parallel pairs and the same training recipe. Interestingly, CompTra behaves differently. As a matter of fact, the traces of CompTra only contain translations of small sentences built by splitting the source, not of the source itself. The translations of these small phrases are unlikely to be better than the ground truth. This explains the performance similarity between standard \textsc{IOFT} and \textsc{IOFT-Max(CompTra)}. \textsc{CoTFT} with CompTra outperforms \textsc{IOFT-Max(CompTra)} and \textsc{IOFT} indicating that \textsc{CoTFT} can be successful without including better translation attempts than the ground truth; partial translations are enough.

\begin{figure}[t]
\begin{center}
    \includegraphics[width=\linewidth]{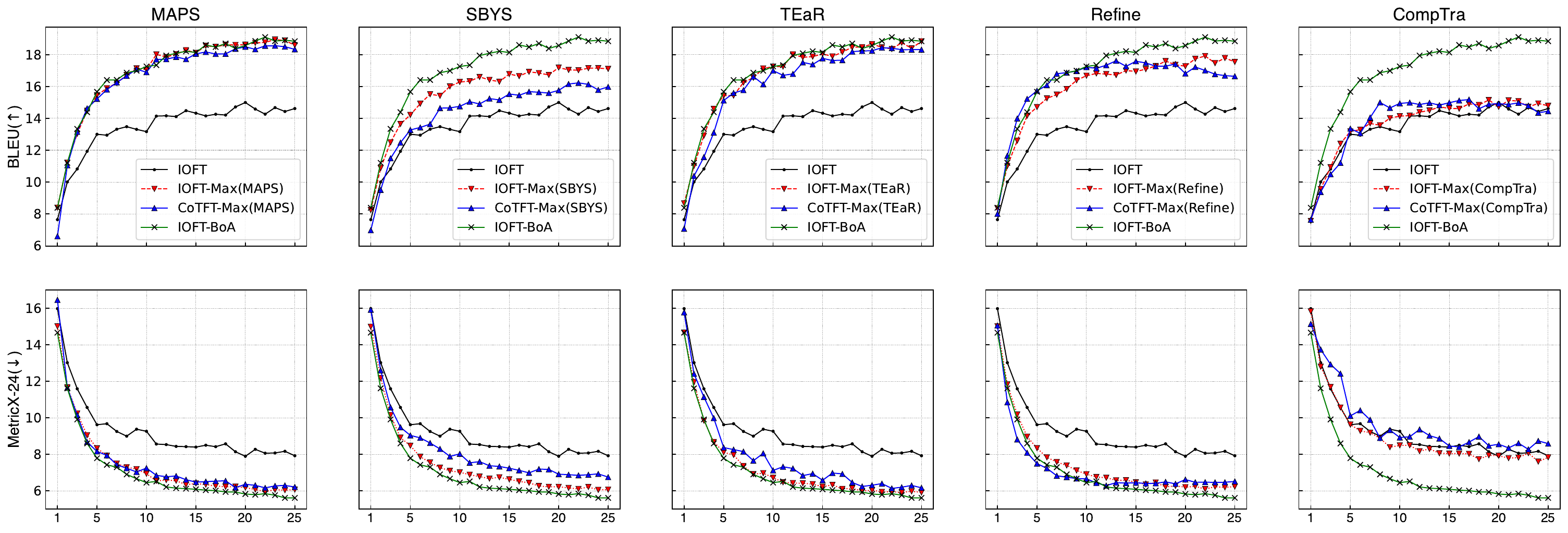}
    \caption{Comparison between \textsc{IOFT} and \textsc{CoTFT} with five different prompting strategies.}
    \label{fig:prompting-v3}
\end{center}
\end{figure}

\paragraph{Second Scenario.} For MAPS, SBYS, TEaR and Self-Refine, \textsc{IOFT-Max} generally works better than \textsc{CoTFT-Max} (Figure~\ref{fig:prompting-v3}). This confirms the previous conclusion on these strategies, i.e.~when the traces provided by the teacher do not contain translation attempts better than the ground truth, they do not help improve the MT performance. \textsc{CoTFT-Max(CompTra)} works slightly better than \textsc{IOFT-Max(CompTra)}, but both underperform \textsc{CoTFT} with CompTra. This reinforces the idea that sentence-translation pairs (related to the sentence considered but smaller and different) can serve as valuable intermediate information for \textsc{CoTFT}.

\textsc{IOFT-BoA} (in green) being consistently above all the curves suggest that the quality of the target translations matters more than traces, and \textsc{IOFT} with better ground truths outperforms \textsc{CoTFT} while being cheaper and faster to train. We obtained the exact same results when we use \texttt{gemma-3-27b-it} as the teacher in Appendix~\ref{appendix:impact_of_intermediate_teacher}.

\section{Discussion and Analysis}

\subsection{Down the rabbit hole of sentence decomposition} \label{subsection:sentence_decomposition}
We further investigate the generation of sentence-translation pairs as intermediate tokens. With CompTra, the pairs are obtained by decomposing the source into multiple phrases \citep{zebaze2025compositionaltranslationnovelllmbased}. We consider three other decomposition strategies: \textit{Paraphrases (P)}, \textit{Syntactic Paraphrases (SP)} and \textit{Hard Expressions (H)}.
\textit{S} asks the teacher to generate five paraphrases of the source. \textit{SP} generates five sentences with the same syntax as the source (grammatical roles, syntactic dependencies etc.) but using different words. Finally, \textit{H} asks the teacher to extract words or expressions it deems difficult to translate.
%
In all cases, the teacher translates the expressions generated after decomposition. For each decomposition strategy (\textit{P}, \textit{SP}, \textit{H}, and CompTra), we compare \textsc{CoTFT} (which uses the teacher’s sentence–translation pairs as reasoning traces{}) with \textsc{IOFT}. We also evaluate \textsc{IOFT-Ext(\textit{strategy})}, which applies \textsc{IOFT} on the original dataset augmented with the generated pairs as additional training samples.
%

\begin{figure}[ht]
  \centering
  \noindent\begin{tabular}{c|c}
    \includegraphics[width=0.75\linewidth]{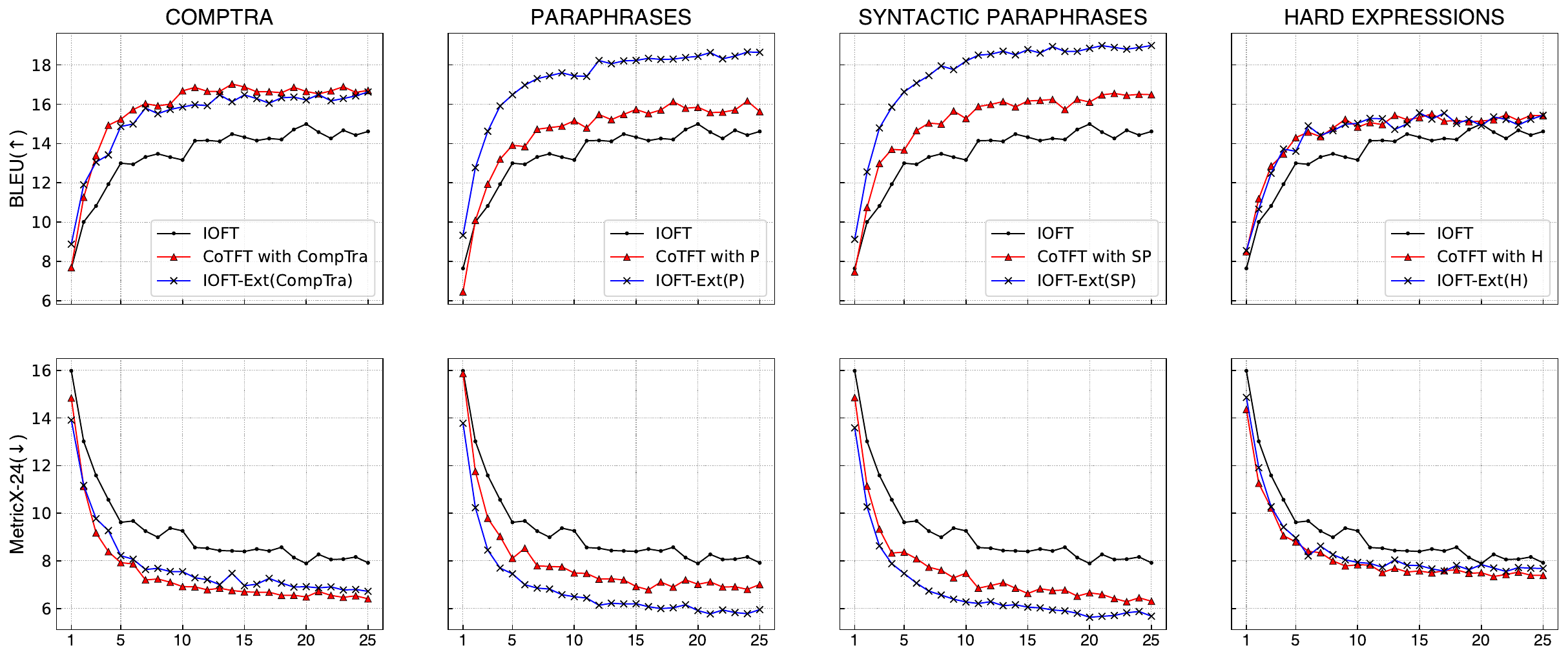} &
    \includegraphics[width=0.21\linewidth]{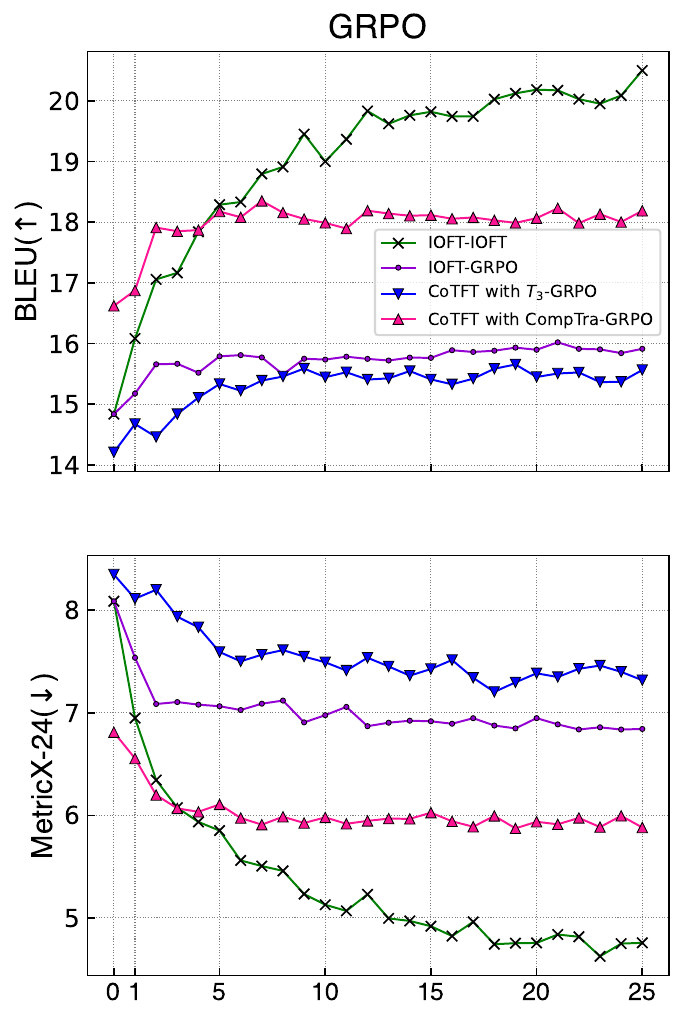}
    \end{tabular}
    \caption{Comparison between \textsc{IOFT} and \textsc{CoTFT} with four sentence decomposition strategies (left) and with GRPO (right).}
    \label{fig:decomp-rl}
\end{figure}

As shown in Figure~\ref{fig:decomp-rl} (8 leftmost panels), \textsc{CoTFT} consistently outperforms \textsc{IOFT} across all decomposition strategies. CompTra and \textit{SP} are the best approaches with \textsc{CoTFT}. \textsc{IOFT-Ext(P)} and \textsc{IOFT-Ext(SP)} result in significant gains over the \textsc{IOFT} baseline (+4 BLEU, -2 MetricX). \textsc{IOFT-Ext} also outperforms \textsc{CoTFT} with \textit{P} and \textit{SP}. However, it has less success with \textit{H} and CompTra. We attribute this to the fact that the pairs generated by \textit{H} and CompTra are shorter and largely overlap with the original training samples, giving fewer gains as additional data compared to entirely new sentences. However, these short phrases (generated via CompTra or \textit{H}) are valuable intermediate information, as \textsc{CoTFT} outperforms \textsc{IOFT} and \textsc{IOFT-Ext} in these scenarios. \textsc{IOFT-Ext(SP)} and \textsc{IOFT-Ext(P)} are the best overall, showing the large impact of the amount of parallel data.

\subsection{Reinforcement Learning After IOFT and CoTFT} \label{section:rl}
Finally, we investigate whether CoTFT improves performance during RL fine-tuning. We consider three setups: \textsc{IOFT}, \textsc{CoTFT} with CompTra and \textsc{CoTFT} with $T_3$ (like in Section~\ref{subsection:cot_distillation}). The final checkpoints (checkpoint-5000) are further fine-tuned with GRPO~\citep{shao2024deepseekmathpushinglimitsmathematical} on a second parallel dataset\footnote{\url{https://hf.co/datasets/almanach/Xhosa}} (on 3 GPUs for 5000 steps, more details in Appendix~\ref{appendix:rl_hyperparameters}). We consider three reward functions: one based on the BLEU and chrF++ scores with the ground truth, a second using COMET-22 (\texttt{wmt22-cometkiwi-da}; \citealp{rei-etal-2022-cometkiwi}) and a last one based on the BLASER-2.0~QE scores between the sources and hypotheses. For \textsc{CoTFT}, we consider an additional format reward to ensure that the models preserve their prior thinking before translating. 
We compare all three RL fine-tunings with \textsc{IOFT} on the second dataset (on 1 GPU for 5000 steps) and report the results in Figure~\ref{fig:decomp-rl} (two rightmost panels). 
%
%
%
The ordering after SFT (\textsc{CoTFT} with $T_3$ $\leq$ \textsc{IOFT} $\leq$ \textsc{CoTFT} with CompTra) remains unchanged after RL, with gains of about +1.3 BLEU and -1.0 MetricX in all setups. Notably, \textsc{CoTFT} still does not outperform \textsc{IOFT}, even with RL. This is consistent with \citeauthor{zheng2025hunyuanmttechnicalreport}'s (\citeyear{zheng2025hunyuanmttechnicalreport}) findings, namely that CoT signals fail to induce meaningful reasoning when the reward is applied only to the final translation. Moreover, unlike mathematics where step-by-step explanations are widely present in pre-training corpora (proofs), it is not the case for translation data. This scarcity of reasoning-like data may explain CoT's limited effectiveness in MT. Finally, we find that continuing SFT (\textsc{IOFT}) on the \textsc{IOFT} checkpoint gives much larger gains (+6 BLEU, -3 MetricX) than GRPO which quickly stagnates, reinforcing that standard \textsc{IOFT} alone can achieve superior MT performance. We report the results of applying GRPO on checkpoints obtained after \textsc{CoTFT} with MAPS, SBYS, TEaR and Self-Refine in Appendix~\ref{appendix:rl}.

\section{Conclusion}
We have explored fine-tuning LLMs to generate intermediate tokens as a method to improve their MT capabilities. Through a broad spectrum of experiments, we find that outputting reasoning traces does not help models to produce better translations (for thinking models and during CoT distillation). We also investigated how traces produced by alternative MT prompting strategies could help and found that parallel pairs can serve as valuable intermediate information. However, ultimately two factors considerably affect the success of MT fine-tuning: the quality of the target translation and the quantity of parallel data. When these factors are on point, standard \textsc{IOFT} goes a long way. These findings generalize two important results in MT: (i)~The inability of CoT prompting to improve over standard IO prompting in zero-shot with standard LLMs \citep{peng-etal-2023-towards, zebaze2025compositionaltranslationnovelllmbased, nguyen-xu-2025-reasoning}, and (ii)~the success of approaches using external resources such as grammars comes from the presence of parallel sentences in the grammar \citep{aycock2025can,marmonier-etal-2025-explicit}. 
CoT (intermediate tokens) provided no benefit when translation attempts (full or partial) were absent, but accounted for all improvements when they were present.
These findings suggest that parallel data is crucial for improving MT, both in the presence and absence of intermediate tokens.

\subsubsection*{Acknowledgments}
This work was partly funded by Rachel Bawden and Benoît Sagot's chairs in the PRAIRIE institute, now PRAIRIE-PSAI, funded by the French national agency ANR, respectively as part of the “Investissements d’avenir” programme under the reference ANR-19-P3IA-0001 and as part of the ``France 2030'' strategy under the reference ANR-23-IACL-0008. It was also partly funded by the French \textit{Agence Nationale de la Recherche} (ANR) under the project TraLaLaM (``ANR-23-IAS1-0006'').This work was granted access to the HPC resources of IDRIS under the allocation 2025-AD011015933 made by GENCI.

\bibliography{custom}
\bibliographystyle{iclr2026_conference}

\appendix

\section{Reproducibility Details} \label{appendix:reproducibility_details}

\subsection{Models, Datasets and Tools} \label{appendix:resources}

In Table~\ref{tab:url}, we list the links to the relevant resources used for the experiments.
\begin{table*}[!ht]
    \centering\small
    \resizebox{\linewidth}{!}{
    \begin{tabular}{l l}
    \toprule
    \multicolumn{2}{c}{\textit{Datasets}} \\
    \midrule
    FLORES-200 & \url{https://huggingface.co/datasets/facebook/flores} \\
    NTREX HF & \url{hhttps://huggingface.co/datasets/mteb/NTREX} \\
    TICO-19 & \url{https://huggingface.co/datasets/gmnlp/tico19} \\
    ToPXGen \texttt{llama-4-scout} \& \texttt{llama-4-scout} & \url{https://hf.co/datasets/almanach/topxgen-llama-4-scout-and-llama-4-scout} \\
    \midrule
    \multicolumn{2}{c}{\textit{Models evaluated}} \\
    \midrule
    \texttt{Qwen3-0.6B} & \url{https://huggingface.co/Qwen/Qwen3-0.6B} \\
    \texttt{Qwen3-1.7B} & \url{https://huggingface.co/Qwen/Qwen3-1.7B} \\
    \texttt{Qwen3-4B}   & \url{https://huggingface.co/Qwen/Qwen3-4B}   \\
    \texttt{Qwen3-8B}   & \url{https://huggingface.co/Qwen/Qwen3-8B}   \\
    \texttt{Qwen3-14B}  & \url{https://huggingface.co/Qwen/Qwen3-14B}  \\
    \texttt{Qwen3-32B}  & \url{https://huggingface.co/Qwen/Qwen3-32B}  \\
    \texttt{DeepSeek-R1-Distill-Qwen-1.5B}  & \url{https://huggingface.co/deepseek-ai/DeepSeek-R1-Distill-Qwen-1.5B} \\
    \texttt{DeepSeek-R1-Distill-Qwen-7B} & \url{https://huggingface.co/deepseek-ai/DeepSeek-R1-Distill-Qwen-7B} \\
    \texttt{DeepSeek-R1-Distill-Llama-8B} & \url{https://huggingface.co/deepseek-ai/DeepSeek-R1-Distill-Llama-8B} \\
    \texttt{DeepSeek-R1-Distill-Qwen-14B} & \url{https://huggingface.co/deepseek-ai/DeepSeek-R1-Distill-Qwen-14B} \\
    \texttt{DeepSeek-R1-Distill-Qwen-32B} & \url{https://huggingface.co/deepseek-ai/DeepSeek-R1-Distill-Qwen-32B} \\
    \texttt{DeepSeek-R1-Distill-Llama-70B} & \url{https://huggingface.co/deepseek-ai/DeepSeek-R1-Distill-Llama-70B} \\
    %
    \texttt{Gemma-3-27B-It} & \url{https://huggingface.co/google/gemma-3-27b-it} \\
    \texttt{Gemma-3-4B-Pt} & \url{https://huggingface.co/google/gemma-3-4b-pt} \\
    \texttt{Gemma-3-1B-Pt} & \url{https://huggingface.co/google/gemma-3-1b-pt} \\
    \texttt{Llama-4-Scout-17B-16E-Instruct} & \url{https://huggingface.co/meta-llama/Llama-4-Scout-17B-16E-Instruct} \\
    \midrule
    \multicolumn{2}{c}{\textit{Other resources}} \\
    \midrule
    MetricX24-Hybrid-XXL & \url{https://huggingface.co/google/metricx-24-hybrid-xxl-v2p6} \\
    XCOMET-XXL & \url{https://huggingface.co/Unbabel/XCOMET-XXL} \\
    FastText & \url{https://huggingface.co/facebook/fasttext-language-identification} \\
    \texttt{vLLM} \citep{kwon2023efficient} & \url{https://github.com/vllm-project/vllm} \\
    \bottomrule
    \end{tabular}
    }
    \caption{Links to datasets, benchmarks and models.}
    \label{tab:url}
\end{table*}

\subsection{Implementation Details} \label{appendix:details}
We use HuggingFace's Transformers library \citep{wolf2020huggingfacestransformersstateoftheartnatural, accelerate}. We adopt the prompt template introduced by \citet{xu2024a}, and compute the loss only on the target (translation or intermediate tokens followed by the translation).  We use the same prompt when evaluating the checkpoints. During \textsc{CoTFT} the target is formatted as \texttt{<think>\textbackslash n\{Intermediate Tokens\}\textbackslash n</think>\textbackslash n\textbackslash nFinal Translation\textbackslash n\{Target translation\}}.

\begin{lstlisting}[frame=single,breaklines=true,basicstyle=\small\ttfamily]
Translate this from English to Hausa:
English: "We now have 4-month-old mice that are non-diabetic that used to be diabetic," he added.
Hausa: 
\end{lstlisting}
For intruction-following models and thinking models we use the following evaluation prompt.

\begin{lstlisting}[frame=single,breaklines=true,basicstyle=\small\ttfamily]
Please write a high-quality Xhosa translation of the following English sentence

"We now have 4-month-old mice that are non-diabetic that used to be diabetic," he added.

Please provide only the translation, nothing more.
\end{lstlisting}

We calculate statistical significance using bootstrap resampling \citep{koehn-2004-statistical} with 300 samples of 500 sentences and a $p$-value threshold of 0.05.

\subsection{CoT Construction Templates} \label{appendix:cot_construction}

We use the six CoT construction templates proposed by \citet{he2025r1t1fullyincentivizingtranslation}. They mimic reasoning strategies for translation commonly adopted by human translators.

T1 is \textbf{Hierarchical Translation}
\begin{lstlisting}[frame=single,breaklines=true,basicstyle=\small\ttfamily]
<think>
1. Analyze the sentence structure and identify the core elements (subject, verb, object).
2. Translate the sentence from the origin language to the target language, focusing on the core elements.
3. Review the translation for basic accuracy and grammatical structure.
4. Identify areas that need further refinement (e.g., word choice, tense, or word order).
5. Modify the translation to improve fluency and coherence, considering the context.
6. Finalize the translation by ensuring it retains the original meaning while improving readability.
</think>
\end{lstlisting}

T2 is \textbf{Triangulating Translation}
\begin{lstlisting}[frame=single,breaklines=true,basicstyle=\small\ttfamily]
<think>
1. Identify basic elements: Break down the sentence into its main components and identify the key subject, verb, and object.
2. Translate to intermediate language: Convert these elements into an intermediate language structure (e.g., simple syntactic rules or function names).
3. Refine back to target language: Translate from the intermediate language back to the target language, adjusting for syntactic norms and idiomatic expressions.
4. Check for accuracy: Ensure that the meaning is preserved in the translated sentence by checking noun-verb agreement and connectors.
5. Adjust word order: Modify word order to ensure that it aligns with the target language's grammatical structure.
6. Final refinement: Review the translation for naturalness, idiomatic use, and overall flow.
</think>
\end{lstlisting}

T3 is \textbf{Back Translation}
\begin{lstlisting}[frame=single,breaklines=true,basicstyle=\small\ttfamily]
<think>
1. Analyze the provided context in the source language.
2. Translate the source text to the target language.
3. Perform back translation from the target language to the source language.
4. Compare the back translation with the original source context.
5. Evaluate whether the meaning of the back translation aligns with the original.
6. If discrepancies are identified, adjust the target language translation to enhance consistency with the original meaning.
7. Finalize the translation by ensuring both forward and back translations accurately align across all languages involved.
</think>
\end{lstlisting}

T4 is \textbf{Context-aware Translation}
\begin{lstlisting}[frame=single,breaklines=true,basicstyle=\small\ttfamily]
<think>
1. Analyze the current sentence, along with the previous sentences, to understand the overall conversation context.
2. Identify key elements like tone, formality, or subject matter based on the ongoing conversation.
3. Translate the sentence while ensuring that the translation is aligned with the tone, style, and subject of the preceding dialogue.
4. If any ambiguity exists in the translation due to context, refine the translation to better fit the conversation flow.
5. Verify that the translation maintains coherence with the larger conversation, ensuring consistency in language and tone.
6. Finalize the translation by cross-checking it with the conversation's context to ensure it feels natural and appropriately aligned.
</think>
\end{lstlisting}

T5 is \textbf{Translation Explanation}
\begin{lstlisting}[frame=single,breaklines=true,basicstyle=\small\ttfamily]
<think>
1. Analyze the source sentence and identify the key elements (verbs, subjects, objects, etc.).
2. Based on these elements, determine the most suitable translation strategy (literal vs. idiomatic).
3. Select the best translation for each word or phrase, considering context and languagespecific structures.
4. Explain the rationale behind choosing specific words or phrases.
5. After completing the initial translation, review each translation decision and explain any adjustments made for fluency or accuracy.
6. Provide a final explanation for the translation choices, discussing any trade-offs made between literal meaning and contextual appropriateness.
</think>
\end{lstlisting}

T6 is \textbf{Structural Transformation}
\begin{lstlisting}[frame=single,breaklines=true,basicstyle=\small\ttfamily]
<think>
1. Analyze the sentence's syntactic structure in the source language (e.g., identify whether it's active or passive).
2. Determine the most appropriate syntactic structure in the target language (e.g., whether it needs to be rephrased from active to passive or vice versa).
3. Adjust the word order and grammatical structure in the target language to match the sentence's meaning, while maintaining clarity.
4. Translate the sentence, ensuring that subject-verb-object relationships and other syntactic elements align with target language norms.
5. After the translation, check the sentence's grammar and overall flow in the target language, making sure it is clear and fluid.
6. If the sentence feels awkward or unnatural, refine the structure by adjusting word choice or reordering components.
</think>
\end{lstlisting}

Given a CoT template, we use the following prompt to obtain a CoT produced by a teacher explaining how to obtain the provided translation of a given source sentence following the strategy corresponding to the template. The CoT produced is in the first-person and can later be used for \textsc{CoTFT}.

\begin{lstlisting}[frame=single,breaklines=true,basicstyle=\small\ttfamily]
Assume that you are a student engaged in translating a sentence from {src} to {tgt}. 
Now you have both the source sentence and the target sentence, and need to analyze how to translate 
from the source sentence to the given target sentence based on the provided Thinking Chain Guide. And
output the chain-of-thought trajectory from source to target sentence.

The {src} statement is as follows:
<Source Sentence>
{sentence}
</Source Sentence>

The {tgt} statement is as follows:
<Target Sentence>
{translation}
</Target Sentence>

You continuously reflect on how to translate the source sentence to the given target sentence
based on the thinking guidance provided.

The given Thinking Chain Guide is as follows:
<Thinking Chain Guide>
{chain_of_thought_template}
</Thinking Chain Guide>

Please refine the entire analysis process into a complete self-reflective description (in the present tense). For self-reflection, you can refer to the following thinking steps: 
directly output the self-reflective description in the <think></think> tags, without any additional descriptions or explanations. 
Each line in the reflective description can be viewed as a reasoning step in the translation process.
\end{lstlisting}

\subsection{Prompting strategies}
\textbf{Step-by-Step Translation (SBYS)}:
\begin{lstlisting}[frame=single,breaklines=true,basicstyle=\small\ttfamily]
{predrafting research}

{draft translation}

Now let's move to the next stage: Post-editing with local refinement.
In this stage, the primary aim is to refine the draft translation by making micro-level improvements that improve the draft's fluency.

Here is a refined version of the translation
{refinement}

Now, we will proofread the refined text for grammar spelling, punctuation, terminology and overall fluency."

Here is the translation after proofreading
{proofreading}

We will further improve it to obtain the final, polished translation.
\end{lstlisting}

\textbf{Multi-Aspect Prompting and Selection (MAPS)}:
\begin{lstlisting}[frame=single,breaklines=true,basicstyle=\small\ttfamily]
Here is a draft translation

1. {zero-shot translation}

Let's write an English sentence related to but different from the input English sentence and translate it into {language}

{demonstrations}

Given this knowledge, we can draft another translation

2. {demonstrations-inspired translation}

Let's extract the keywords in the provided English sentence, and then translate these keywords into {language}

{keywords}

Given this knowledge, we can draft another translation

3. {keywords-inspired translation}

Let's use a few words to describe the topics of the provided English sentence

{topics}

Given this knowledge, we can draft another translation

4. {topics-inspired translation}

We will choose the best of these translations and further improve it to obtain the final, polished translation.
\end{lstlisting}

\textbf{Self-Refine}
\begin{lstlisting}[frame=single,breaklines=true,basicstyle=\small\ttfamily]
Here is a draft translation

1. {draft translation}

Let's improve it and write a better translation

2. {refinement 1}

Let's further improve it and write a better translation

3. {refinement 2}

Let's improve it one last time and write a better translation

4. {refinement 3}

We will choose the best of these translations and further improve it to obtain the final, polished translation.
\end{lstlisting}

\textbf{Translate, Estimate and Refine (TEaR)}
\begin{lstlisting}[frame=single,breaklines=true,basicstyle=\small\ttfamily]
Here is a draft translation

1. {draft translation}

Let's identify errors and assess the quality of the draft translation.
The categories of errors are accuracy (addition, mistranslation, omission, untranslated text), fluency (character encoding, grammar, inconsistency, punctuation, register, spelling), locale convention (currency, date, name, telephone, or time format) style (awkward), terminology (inappropriate for context, inconsistent use), non translation, other, or no-error.
Each error is classified as one of three categories: critical, major, and minor. Critical errors inhibit comprehension of the text. Major errors disrupt the flow, but what the text is trying to say is still understandable. Minor errors are technical errors but do not disrupt the flow or hinder comprehension.


Here are the MQM annotations of the draft:
{MQM annotations}

Upon reviewing the translation and error information, we can refine the draft and obtain a better translation

2. {refinement}

We will further improve it to obtain the final, polished translation."
\end{lstlisting}

\textbf{Compositional Translation (CompTra)}
\begin{lstlisting}[frame=single,breaklines=true,basicstyle=\small\ttfamily]
1. English Sentence
{}
Xhosa Translation
{}

2. English Sentence
{}
Xhosa Translation
{}

3. English Sentence
{}
Xhosa Translation
{}
\end{lstlisting}

\subsection{RL Training Hyperparameters} \label{appendix:rl_hyperparameters}
For GRPO, we use the Hugging Face TRL library \citep{vonwerra2022trl}. Training is conducted on four H100 GPUs, with one dedicated to model deployment for reward computation. We set a per-device batch size of 4 with 4 gradient accumulation steps, for a total of 5000 steps including 100 warmup steps. Hyperparameters include a beta value of 0.02, a maximum gradient norm of 1.0, and a temperature of 1.0. For generation, we sample 12 outputs per prompt with an effective batch size of 48. We apply LoRA \citep{hu2022lora}, fine-tuning the \texttt{q\_proj}, \texttt{k\_proj}, \texttt{v\_proj}, \texttt{o\_proj}, \texttt{gate\_proj}, \texttt{up\_proj}, and \texttt{down\_proj} modules with rank $r=32$, scaling factor $\alpha=64$, and dropout rate 0.05.

\section{Additional Experiments} \label{appendix:additional_experiments}

\begin{table*}[ht]
\vskip 0.15in
\small
\begin{center}
\resizebox{\textwidth}{!}{
\begin{tabular}{lrrrrrrrrrrrrrr}
\toprule
\multirow{2}{*}{Models}  & \multicolumn{2}{c}{Czech} & & \multicolumn{2}{c}{Finnish} & & \multicolumn{2}{c}{French} & & \multicolumn{2}{c}{German} & & \multicolumn{2}{c}{Japanese}\\
\cmidrule{2-3} \cmidrule{5-6} \cmidrule{8-9} \cmidrule{11-12} \cmidrule{14-15}
{} & {chrF++} & {XCOMET} &  & {chrF++} & {XCOMET} & & {chrF++} & {XCOMET} & & {chrF++} & {XCOMET} & & {chrF++} & {XCOMET}\\
\midrule
\textsc{Qwen3-0.6B}           & \bf 19.43 & 16.52 &  & 16.36 & 13.56 & & \bf 47.19 & 53.77 & & \bf 39.34 & 71.05 & & \bf 14.11 & 44.66  \\
\textit{\quad w/o Thinking}   & 18.53 & \bf 55.22 &  & \bf 18.12 & \bf 23.62 & & 45.94 & \bf 54.82 & & 37.75 & \bf 74.96 & & 11.35 & \bf 53.73  \\
\midrule
\textsc{Qwen3-1.7B}           & 33.89 & \bf 39.91 &  & 26.45 & 20.00 & & 57.40 & 80.58 & & 49.52 & 88.42 & & \bf 21.39 & \bf 69.82 \\
\textit{\quad w/o Thinking}   & \bf 35.27 & 38.62 &  & \bf 29.54 & \bf 22.68 & & \bf 57.68 & \bf 80.67 & & \bf 49.89 & \bf 89.58 & & 20.81 & 67.83 \\
\midrule
\textsc{Qwen3-4B}             & 43.43 & \bf 68.22 &  & 36.99 & \bf 45.99 & & \bf 62.49 & \bf 89.14 & & \bf 55.13 & 93.48 & & 24.68 & \bf 80.26 \\
\textit{\quad w/o Thinking}   & \bf 43.92 & 65.45 &  & \bf 38.11 & 42.54 & & 62.26 & 88.23 & & \bf 55.07 & \bf 94.25 & & \bf 26.53 & 79.23  \\
\midrule
\textsc{Qwen3-8B}             & 48.49 & \bf 80.32 &  & 43.15 & \bf 65.78 & & 64.68 & \bf 92.19 & & \bf 58.30 & 95.94 & & 27.24 & 86.18 \\
\textit{\quad w/o Thinking}   & \bf 48.70 & 77.64 &  & \bf 43.44 & 62.60 & & \bf 65.16 & 91.89 & & 58.21 & \bf 96.38 & & \bf 28.50 & \bf 86.60  \\
\midrule
\textsc{Qwen3-14B}            & \bf 51.57 & \bf 86.91 &  & 46.26 & \bf 76.27 & & 66.41 & 92.99 & & \bf 60.52 & 96.89 & & 29.88 & \bf 89.88  \\
\textit{\quad w/o Thinking}   & 51.29 & 84.65 &  & \bf 46.78 & 74.75 & & \bf 66.91 & \bf 93.57 & & 60.18 & \bf 97.37 & & \bf 30.54 & 89.76  \\
\midrule
\textsc{Qwen3-32B}            & \bf 51.45 & \bf 86.59 &  & 46.47 & \bf 77.18 & & \bf 65.99 & 92.79 & & \bf 60.35 & 96.27 & & 29.10 & 89.38  \\
\textit{\quad w/o Thinking}   & 50.85 & \bf 86.48 &  & \bf 47.40 & \bf 77.13 & & \bf 65.98 & \bf 93.81 & & \bf 60.38 & 9\bf 7.60 & & \bf 30.89 & \bf 89.97  \\
\end{tabular}
}
\resizebox{\textwidth}{!}{
\begin{tabular}{lrrrrrrrrrrrrrr}
\toprule
\multirow{2}{*}{Models}  & \multicolumn{2}{c}{Kazakh} & & \multicolumn{2}{c}{Lithuanian} & & \multicolumn{2}{c}{Portuguese} & & \multicolumn{2}{c}{Spanish} & & \multicolumn{2}{c}{Turkish}\\
\cmidrule{2-3} \cmidrule{5-6} \cmidrule{8-9} \cmidrule{11-12} \cmidrule{14-15}
{} & {chrF++} & {XCOMET} &  & {chrF++} & {XCOMET} & & {chrF++} & {XCOMET} & & {chrF++} & {XCOMET} & & {chrF++} & {XCOMET}\\
\midrule
\textsc{Qwen3-0.6B}           & 4.14 & 15.77 &  & 10.73 & 14.19 & & \bf 48.05 & 67.21 & & \bf 41.70 & 70.45 & & 22.64 & 18.39  \\
\textit{\quad w/o Thinking}   & \bf 6.78 & \bf 20.31 &  & \bf 12.97 & \bf 16.91 & & 44.73 & \bf 68.36 & & 40.29 & \bf 72.40 & & \bf 23.77 & \bf 22.97  \\
\midrule
\textsc{Qwen3-1.7B}           & 7.51 & 15.09 &  & 24.83 & 18.80 & & \bf 58.97 & \bf 87.43 & & 48.25 & \bf 87.84 & & 34.08 & 39.47 \\
\textit{\quad w/o Thinking}   & \bf 10.84 & \bf 15.62 &  & \bf 26.11 & \bf 19.52 & & \bf 58.99 & 87.07 & & \bf 48.61 & 87.55 & & \bf 35.89 & \bf 43.02 \\
\midrule
\textsc{Qwen3-4B}             & \bf 26.54 & \bf 25.20 &  & 36.26 & \bf 43.79 & & \bf 63.93 & \bf 92.69 & & 50.66 & \bf 92.23 & & 42.97 & \bf 66.46  \\
\textit{\quad w/o Thinking}   & \bf 26.51 & 23.78 &  & \bf 36.60 & 42.88 & & 63.70 & 92.17 & & \bf 51.07 & \bf 92.20 & & \bf 43.52 & 65.43  \\
\midrule
\textsc{Qwen3-8B}             & \bf 34.64 & \bf 40.48 &  & 41.54 & \bf 61.88 & & 65.66 & 94.16 & & 52.45 & 94.13 & & \bf 47.86 & \bf 77.71  \\
\textit{\quad w/o Thinking}   & 33.86 & 36.52 &  & \bf 42.02 & 60.84 & & \bf 65.77 & \bf 94.60 & & \bf 52.55 & \bf 94.62 & & \bf 47.90 & \bf 77.68  \\
\midrule
\textsc{Qwen3-14B}            & \bf 39.53 & \bf 53.44 &  & 45.08 & \bf 73.81 & & 66.70 & 95.12 & & 53.29 & 94.64 & & \bf 50.97 & \bf 83.68  \\
\textit{\quad w/o Thinking}   & 38.95 & 50.77 &  & \bf 45.82 & \bf 73.72 & & \bf 67.55 & \bf 95.31 & & \bf 53.60  & \bf 95.25 & & 50.42 & 83.35  \\
\midrule
\textsc{Qwen3-32B}            & 38.34 & \bf 53.66 &  & 45.92 & 75.60 & & 66.71 & 94.93 & & 53.19 & 95.07 & & \bf 51.29 & \bf 84.03  \\
\textit{\quad w/o Thinking}   & \bf 39.22 & 52.05 &  & \bf 46.68 & \bf 76.89 & & \bf 67.64 & \bf 95.77 & & \bf 53.57 & \bf 95.32 & & 50.60 & 83.08  \\
\bottomrule
\end{tabular}
}
\end{center}
\caption{chrF++ and XCOMET scores for 10 English $\rightarrow$ X directions from FLORES~200\iffalse~\citep{goyal-etal-2022-flores, nllb2022}\fi. Best results are highlighted in bold.}
\label{tab:zs_hrl_comet}
\end{table*} 

\begin{table*}[ht]
\vskip 0.15in
\small
\begin{center}
\resizebox{\textwidth}{!}{
\begin{tabular}{lrrrrrrrrrrrrrr}
\toprule
\multirow{2}{*}{Models}  & \multicolumn{2}{c}{Czech} & & \multicolumn{2}{c}{Finnish} & & \multicolumn{2}{c}{French} & & \multicolumn{2}{c}{German} & & \multicolumn{2}{c}{Japanese}\\
\cmidrule{2-3} \cmidrule{5-6} \cmidrule{8-9} \cmidrule{11-12} \cmidrule{14-15}
{} & {BLEU} & {MetricX} &  & {BLEU} & {MetricX} & & {BLEU} & {MetricX} & & {BLEU} & {MetricX} & & {BLEU} & {MetricX}\\
\midrule
\textsc{DeepSeek-R1-Distill-Qwen-14B}             & 22.78 & 9.91 & & \bf 10.55 & 17.40 & & \bf 46.13 & 2.87 & & \bf 34.54 & 2.49 & & 21.04 & 4.85 \\
\textit{\quad w/o Thinking}                       & \bf 23.04 & \bf 9.50 & & 10.00 & \bf 17.13 & & 45.42 & \bf 2.54 & & 30.26 & \bf 2.40 & & \bf 22.04 & \bf 4.45 \\
\midrule
\textsc{DeepSeek-R1-Distill-Qwen-32B}             & 28.49 & 7.12 & & 15.61 & 13.06 & & 48.68 & 2.78 & & 38.25 & 1.82 & & 24.11 & 4.56 \\
\textit{\quad w/o Thinking}                       & \bf 29.61 & \bf 6.22 & & \bf 16.12 & \bf 12.52 & & \bf 49.35 & \bf 2.16 & & \bf 38.95 & \bf 1.44 & & \bf 26.57 & \bf 3.89  \\
\midrule
\textsc{DeepSeek-R1-Distill-Llama-70B}            & 37.31 & 4.19 &  & 29.90 & 5.35 & & \bf 52.34 & 2.22 & & 43.39 & 1.16 & & 25.61 & 4.08 \\
\textit{\quad w/o Thinking}                       & \bf 38.47 & \bf 3.52 &  & \bf 30.96 & \bf 4.38 & & \bf 52.34 & \bf 1.88 & & \bf 44.78 & \bf 0.91 & & \bf 27.94 & \bf 3.62 \\
\end{tabular}
}
\resizebox{\textwidth}{!}{
\begin{tabular}{lrrrrrrrrrrrrrr}
\toprule
\multirow{2}{*}{Models}  & \multicolumn{2}{c}{Kazakh} & & \multicolumn{2}{c}{Lithuanian} & & \multicolumn{2}{c}{Portuguese} & & \multicolumn{2}{c}{Spanish} & & \multicolumn{2}{c}{Turkish}\\
\cmidrule{2-3} \cmidrule{5-6} \cmidrule{8-9} \cmidrule{11-12} \cmidrule{14-15}
{} & {BLEU} & {MetricX} &  & {BLEU} & {MetricX} & & {BLEU} & {MetricX} & & {BLEU} & {MetricX} & & {BLEU} & {MetricX}\\
\midrule
\textsc{DeepSeek-R1-Distill-Qwen-14B}             & 2.38 & 13.25 & & \bf 7.87 & 20.02 & & 46.28 & 2.86 & & 29.48 & 2.56 & & \bf 17.99 & 12.79 \\
\textit{\quad w/o Thinking}                       & \bf 2.50 & \bf 6.49 & & \bf 7.21 & \bf 18.79 & & \bf 47.11 & \bf 2.52 & & \bf 30.52 & \bf 2.13 & & 17.78 & \bf 11.70 \\
\midrule
\textsc{DeepSeek-R1-Distill-Qwen-32B}             & \bf 5.66 & 18.37 & & 13.24 & 16.09 & & 49.17 & 2.55 & & 30.62 & 2.26 & & 24.52 & 9.36 \\
\textit{\quad w/o Thinking}                       & 4.95 & \bf 16.19 & & \bf 14.02 & \bf 14.92 & & \bf 50.53 & \bf 2.02 & & \bf 31.67 & \bf 1.74 & & \bf 24.96 & \bf 7.88  \\
\midrule
\textsc{DeepSeek-R1-Distill-Llama-70B}            & 21.39 & 8.14 &  & 25.63 & 8.22 & & 52.27 & 2.09 & & 32.68 & 1.78 & & 33.42 & 5.46 \\
\textit{\quad w/o Thinking}                       & \bf 21.56 & \bf 6.76 &  & \bf 27.01 & \bf 7.33 & & \bf 52.64 & \bf 1.75 & & \bf 33.54 & \bf 1.55 & & \bf 34.77 & \bf 5.00 \\
\bottomrule
\end{tabular}
}
\end{center}
\caption{BLEU and MetricX scores for 10 English $\rightarrow$ X directions from FLORES~200\iffalse~\citep{goyal-etal-2022-flores, nllb2022}\fi. Best results are highlighted in bold.}
\label{tab:dzs_hrl}
\end{table*} 

\begin{table*}[ht]
\vskip 0.15in
\small
\begin{center}
\resizebox{\textwidth}{!}{
\begin{tabular}{lrrrrrrrrrrrrrr}
\toprule
\multirow{2}{*}{Models}  & \multicolumn{2}{c}{Czech} & & \multicolumn{2}{c}{Finnish} & & \multicolumn{2}{c}{French} & & \multicolumn{2}{c}{German} & & \multicolumn{2}{c}{Japanese}\\
\cmidrule{2-3} \cmidrule{5-6} \cmidrule{8-9} \cmidrule{11-12} \cmidrule{14-15}
{} & {BLEU} & {MetricX} &  & {BLEU} & {MetricX} & & {BLEU} & {MetricX} & & {BLEU} & {MetricX} & & {BLEU} & {MetricX}\\
\midrule

\textsc{Qwen3-0.6B}           & \bf 5.65 & 22.53 & & \bf 3.21 & \bf 23.13 & & \bf 24.49 & \bf 9.51 & & \bf 16.11 & \bf 10.15 & & \bf 8.08 & \bf 9.29 \\
\textit{\quad w/o Thinking}   & 5.31 & \bf 22.49 & & 2.69 & 23.40 & & 21.48 & 10.28 & & 14.51 & 10.43 & & 6.09 & 11.84 \\
\midrule
\textsc{Qwen3-1.7B}           & 15.92 & 14.68 & & 7.86 & 19.18 & & 38.28 & 4.69 & & \bf 28.41 & 4.30 & & \bf 16.46 & \bf 5.85 \\
\textit{\quad w/o Thinking}   & \bf 16.26 & \bf 14.48 & & \bf 8.50 & \bf 18.71 & & \bf 38.52 & \bf 4.54 & & 28.19 & \bf 4.05 & & 15.29 & \bf 5.89 \\
\midrule
\textsc{Qwen3-4B}             & 24.43 & 9.18 & & 13.96 & 13.88 & & 44.47 & 3.66 & & 34.20 & 3.14 & & 20.68 & 5.22 \\
\textit{\quad w/o Thinking}   & \bf 25.77 & \bf 8.74 & & \bf 15.33 & \bf 13.39 & & \bf 44.99 & \bf 3.25 & & \bf 34.79 & \bf 2.54 & & \bf 21.36 & \bf 4.79 \\
\midrule
\textsc{Qwen3-8B}             & 30.11 & 6.65 & & 19.37 & 10.21 & & 48.89 & 2.89 & & 38.88 & 1.92 & & \bf 24.88 & 4.28 \\
\textit{\quad w/o Thinking}   & \bf 30.36 & \bf 6.59 & & \bf 19.70 & \bf 9.89 & & \bf 49.18 & \bf 2.58 & & \bf 39.16 & \bf 1.59 & & 24.67 & \bf 4.10\\
\midrule
\textsc{Qwen3-14B}            & \bf 34.44 & 5.11 & & 23.51 & 7.59 & & 51.23 & 2.37 & & \bf 41.83 & 1.39 & & \bf 27.93 & 3.86 \\
\textit{\quad w/o Thinking}   & 34.01 & \bf 4.93 & & \bf 23.89 & \bf 7.43 & & \bf 51.45 & \bf 2.13 & & \bf 41.76 & \bf 1.17 & & \bf 28.00 & \bf 3.63 \\
\midrule
\textsc{Qwen3-32B}            & 15.69 & 15.01 & & 11.19 & 16.04 & & 29.14 & 11.67 & & 22.99 & 11.00 & & 16.14 & 12.00 \\
\textit{\quad w/o Thinking}   & \bf 34.34 & \bf 4.64 & & \bf 24.91 & \bf 6.72 & & \bf 50.35 & \bf 1.99 & & \bf 42.58 & \bf 1.09 & & \bf 27.86 & \bf 3.53  \\
\end{tabular}
}
\resizebox{\textwidth}{!}{
\begin{tabular}{lrrrrrrrrrrrrrr}
\toprule
\multirow{2}{*}{Models}  & \multicolumn{2}{c}{Kazakh} & & \multicolumn{2}{c}{Lithuanian} & & \multicolumn{2}{c}{Portuguese} & & \multicolumn{2}{c}{Spanish} & & \multicolumn{2}{c}{Turkish}\\
\cmidrule{2-3} \cmidrule{5-6} \cmidrule{8-9} \cmidrule{11-12} \cmidrule{14-15}
{} & {BLEU} & {MetricX} &  & {BLEU} & {MetricX} & & {BLEU} & {MetricX} & & {BLEU} & {MetricX} & & {BLEU} & {MetricX}\\
\midrule
\textsc{Qwen3-0.6B}           & 0.93 & \bf 23.89 & & \bf 2.21 & \bf 24.04 & & \bf 26.00 & \bf 9.44 & & \bf 17.51 & \bf 8.38 & & \bf 6.46 & 21.24 \\
\textit{\quad w/o Thinking}   & \bf 1.31 & \bf 23.85 & & 2.11 & 24.12 & & 21.99 & 11.08 & & 15.81 & 9.11 & & 5.43 & \bf 21.16  \\
\midrule
\textsc{Qwen3-1.7B}           & 2.03 & 22.03 & & 6.80 & \bf 20.51 & & \bf 39.27 & 4.41 & & \bf 25.76 & 3.85 & & 14.10 & 14.36  \\
\textit{\quad w/o Thinking}   & \bf 3.24 & \bf 21.58 & & \bf 7.69 & 20.65 & & \bf 39.22 & \bf 4.23 & & 25.63 & \bf 3.76 & & \bf 14.71 & \bf 13.41  \\
\midrule
\textsc{Qwen3-4B}             & 10.1 & 15.26 & & 13.64 & 15.10 & & 45.61 & 3.44 & & 28.43 & 3.14 & & 21.32 & 10.28  \\
\textit{\quad w/o Thinking}   & \bf 11.0 & \bf 14.50 & & \bf 14.62 & \bf 14.44 & & \bf 45.93 & \bf 3.16 & & \bf 29.12 & \bf 2.69 & & \bf 21.95 & \bf 9.63 \\
\midrule
\textsc{Qwen3-8B}             & 15.64 & 11.07 & & 19.01 & 10.97 & & 49.03 & 2.64 & & 30.87 & 2.30 & & \bf 27.01 & 7.34 \\
\textit{\quad w/o Thinking}   & \bf 15.77 & \bf 10.80 & & \bf 19.51 & \bf 10.67 & & \bf 49.44 & \bf 2.47 & & \bf 31.19 & \bf 2.09 & & \bf 26.98 & \bf 6.82\\
\midrule
\textsc{Qwen3-14B}            & \bf 20.35 & 8.30 & & 23.68 & 8.41 & & 50.93 & 2.28 & & 32.38 & 1.94 & & \bf 30.99 & 6.08  \\
\textit{\quad w/o Thinking}   & 19.40 & \bf 7.97 & & \bf 24.59 & \bf 7.86 & & \bf 51.37 & \bf 2.10 & & \bf 32.67 & \bf 1.72 & & 29.97 & \bf 5.66 \\
\midrule
\textsc{Qwen3-32B}            & 10.85 & 15.88 & & 11.89 & 16.26 & & 29.30 & 11.61 & & 17.69 & 11.53 & & 16.08 & 14.69 \\
\textit{\quad w/o Thinking}   & \bf 20.63 & \bf 7.61 & & \bf 25.28 & \bf 7.12 & & \bf 50.89 & \bf 1.95 & & \bf 33.05 & \bf 1.58 & & \bf 30.97 & \bf 5.26\\
\bottomrule
\end{tabular}
}
\end{center}
\caption{5-shot BLEU and MetricX scores for 10 English $\rightarrow$ X directions from FLORES~200\iffalse~\citep{goyal-etal-2022-flores, nllb2022}\fi. Best results are highlighted in bold.}
\label{tab:fs_hrl}
\end{table*} 

\begin{table*}[ht]
\vskip 0.15in
\small
\begin{center}
\resizebox{\textwidth}{!}{
\begin{tabular}{lrrrrrrrrrrrrrr}
\toprule
\multirow{2}{*}{Models}  & \multicolumn{2}{c}{Czech} & & \multicolumn{2}{c}{Finnish} & & \multicolumn{2}{c}{French} & & \multicolumn{2}{c}{German} & & \multicolumn{2}{c}{Japanese}\\
\cmidrule{2-3} \cmidrule{5-6} \cmidrule{8-9} \cmidrule{11-12} \cmidrule{14-15}
{} & {BLEU} & {MetricX} &  & {BLEU} & {MetricX} & & {BLEU} & {MetricX} & & {BLEU} & {MetricX} & & {BLEU} & {MetricX}\\
\midrule
\textsc{Qwen3-0.6B}           & \bf 23.33 & \bf 7.29 & & \bf 11.40 & 13.01 & & \bf 35.95 & \bf 3.32 & & \bf 33.10 & \bf 4.12 & & \bf 15.47 & \bf 5.96 \\
\textit{\quad w/o Thinking}   & 22.05 & 7.42 & &  9.30 & \bf 11.40 & & 35.02 & \bf 3.31 & & 31.26 & 4.24 & & 14.88 & 6.38 \\
\midrule
\textsc{Qwen3-1.7B}           & \bf 34.90 & \bf 3.33 & & \bf 23.42 & \bf 6.25 & & 41.40 & \bf 1.99 & & 40.50 & \bf 2.29 & & \bf 23.67 & \bf 3.03 \\
\textit{\quad w/o Thinking}   & 34.23 & 3.54 & & 22.62 & \bf 6.25 & & \bf 41.89 & 2.03 & & \bf 40.94 & \bf 2.31 & & 22.66 & 3.41 \\
\midrule
\textsc{Qwen3-4B}             & 38.76 & \bf 2.15 & & 30.57 & 3.68 & & 44.54 & \bf 1.59 & & 44.43 & \bf 1.59 & & 27.05 & \bf 2.26 \\
\textit{\quad w/o Thinking}   & \bf 39.06 & 2.26 & & \bf 30.70 & \bf 3.47 & & \bf 45.59 & \bf 1.61 & & \bf 44.83 & 1.66 & & \bf 27.32 & 2.40 \\
\midrule
\textsc{Qwen3-8B}             & 40.02 & \bf 1.80 & & 33.28 & 2.52 & & 45.80 & \bf 1.42 & & 45.14 & 1.50 & & 27.98 & \bf 1.92 \\
\textit{\quad w/o Thinking}   & \bf 40.77 & 1.86 & & \bf 33.49 & \bf 2.46 & & \bf 46.54 & \bf 1.43 & & \bf 45.46 & \bf 1.42 & & \bf 28.95 & \bf 1.93 \\
\midrule
\textsc{Qwen3-14B}            & 41.91 & 1.60 & & 35.45 & 2.12 & & 46.70 & 1.37 & & 46.28 & 1.36 & & 29.20 & \bf 1.75 \\
\textit{\quad w/o Thinking}   & \bf 43.28 & \bf 1.57 & & \bf 36.31 & \bf 2.00 & & \bf 48.64 & \bf 1.32 & & \bf 47.49 & \bf 1.32 & & \bf 30.43 & \bf 1.77 \\
\midrule
\textsc{Qwen3-32B}            & 43.19 & 1.47 & & 37.01 & 1.84 & & 47.51 & \bf 1.27 & & 46.77 & 1.29 & & 29.99 & 1.68 \\
\textit{\quad w/o Thinking}   & \bf 44.23 & \bf 1.42 & & \bf 37.84 & \bf 1.77 & & \bf 48.72 & \bf 1.26 & & \bf 47.52 & \bf 1.26 & & \bf 30.88 & \bf 1.70 \\
\end{tabular}
}
\resizebox{\textwidth}{!}{
\begin{tabular}{lrrrrrrrrrrrrrr}
\toprule
\multirow{2}{*}{Models}  & \multicolumn{2}{c}{Kazakh} & & \multicolumn{2}{c}{Lithuanian} & & \multicolumn{2}{c}{Portuguese} & & \multicolumn{2}{c}{Spanish} & & \multicolumn{2}{c}{Turkish}\\
\cmidrule{2-3} \cmidrule{5-6} \cmidrule{8-9} \cmidrule{11-12} \cmidrule{14-15}
{} & {BLEU} & {MetricX} &  & {BLEU} & {MetricX} & & {BLEU} & {MetricX} & & {BLEU} & {MetricX} & & {BLEU} & {MetricX}\\
\midrule
\textsc{Qwen3-0.6B}           & \bf 6.94 & 15.07 & & \bf 10.21 & 13.71 & & \bf 39.22 & \bf 3.59 & & \bf 25.92 & \bf 3.92 & & \bf 16.12 & \bf 9.21 \\
\textit{\quad w/o Thinking}   & 5.92 & \bf 15.67 & &  4.36 &  \bf 7.41 & & 37.72 & 3.70 & & 25.64 & 3.99 & & 13.83 & 9.34 \\
\midrule
\textsc{Qwen3-1.7B}           & \bf 17.90 & \bf 8.48 & & \bf 23.61 & \bf 6.26 & & 45.63 & \bf 2.15 & & 29.79 & \bf 2.43 & & \bf 29.31 & \bf 4.39 \\
\textit{\quad w/o Thinking}   & 16.20 & 9.17 & & 22.23 & 6.54 & & \bf 46.32 & 2.31 & & \bf 30.49 & 2.46 & & 27.65 & 4.53 \\
\midrule
\textsc{Qwen3-4B}             & \bf 25.64 & \bf 5.02 & & \bf 29.64 & \bf 3.75 & & 49.12 & \bf 1.79 & & 32.59 & \bf 1.95 & & \bf 34.96 & \bf 2.77 \\
\textit{\quad w/o Thinking}   & 24.20 & 5.31 & & 29.37 & 3.83 & & \bf 50.16 & 1.84 & & \bf 33.55 & 2.02 & & 34.69 & 2.88 \\
\midrule
\textsc{Qwen3-8B}             & \bf 29.31 & \bf 3.80 & & 33.01 & \bf 2.76 & & 50.04 & \bf 1.54 & & 32.89 & \bf 1.75 & & \bf 37.88 & \bf 2.15 \\
\textit{\quad w/o Thinking}   & 28.91 & 3.98 & & \bf 33.28 & 2.81 & & \bf 50.89 & \bf 1.56 & & \bf 33.80 & \bf 1.78 & & 37.77 & 2.24 \\
\midrule
\textsc{Qwen3-14B}            & 32.01 & \bf 3.06 & & 34.31 & 2.48 & & 51.22 & \bf 1.48 & & 34.06 & \bf 1.63 & & 39.52 & \bf 1.88 \\
\textit{\quad w/o Thinking}   & \bf 32.17 & \bf 3.08 & & \bf 35.15 & \bf 2.43 & & \bf 53.28 & \bf 1.49 & & \bf 35.50 & \bf 1.63 & & \bf 40.18 & \bf 1.86 \\
\midrule
\textsc{Qwen3-32B}            & 33.33 & 3.27 & & 36.10 & \bf 2.06 & & 51.69 & 1.44 & & 34.49 & 1.57 & & 40.91 & 1.81 \\
\textit{\quad w/o Thinking}   & \bf 34.03 & \bf 2.77 & & \bf 37.15 & 2.14 & & \bf 53.66 & \bf 1.39 & & \bf 35.85 & \bf 1.54 & & \bf 41.87 & \bf 1.72 \\
\bottomrule
\end{tabular}
}
\end{center}
\caption{BLEU and MetricX scores for 10 X $\rightarrow$ English directions from FLORES~200\iffalse~\citep{goyal-etal-2022-flores, nllb2022}\fi. Best results are highlighted in bold.}
\label{tab:zs_hrl_reverse}
\end{table*} 

\subsection{Benchmarking LRMs at Scale: To Think, or Not To Think?} \label{appendix:preliminary_experiments}
In this section we further investigate the impact of thinking tokens when benchmarking LRMs. In Table~\ref{tab:zs_hrl_comet} we report the chrF++\footnote{nrefs:1$|$case:mixed$|$eff:yes$|$nc:6$|$nw:2$|$space:no$|$version:2.4.2}~\citep{popovic-2015-chrf, popovic:2017:WMT} and \texttt{XCOMET-XXL} \citep{guerreiro-etal-2024-xcomet} scores in the same setup as Table~\ref{tab:zs_hrl}. They tell the same story as BLEU and MetricX. Outputting thinking tokens only marginally helps; the gains are not consistent and when they occur they are small. This questions the necessity of an LRM to think before doing MT, all the more so that thinking is considerably more expensive than straight up answering. It is worth noting that ``small'' models (Qwen-0.6B, Qwen-1.7B and Qwen-4B) often generate answers in English or Chinese when they struggle with the target language (e.g.,~Czech, Finnish, Kazakh, Lithuanian etc.) resulting in artificially better neural scores. The thinking mode particularly helps in such scenarios because it allows the model to remember that it should write an answer in a different language than what it is ``used'' to generating in. When the models are big enough (typically $\geq$ 8B), thereby solving this incorrect language issue, thinking does not result in any gains. Moreover, we run additional experiments with more DeepSeek-R1-Distill models (see Table~\ref{tab:dzs_hrl}) and again we observe a similar pattern, ``no thinking'' consistently outperforms thinking.
We run additional experiments when translating in the 5-shot setting, retrieving demonstrations from the FLORES 200 dev test with bm25s \citep{bm25s} following \citet{zebaze-etal-2025-context}. As shown in Table~\ref{tab:fs_hrl}, providing demonstrations does not help the thinking mode take performance to an upper level. Similarly, it does not help when translating into English, as reported in Table~\ref{tab:zs_hrl_reverse}.

\subsection{Results on NTREX-128 and TICO-19} \label{appendix:ntrex_and_tico}
In this section, we evaluate the models on 2 additional benchmarks:
\begin{itemize}[noitemsep, topsep=0pt, leftmargin=*]
    \item \textbf{NTREX~128}~\citep{barrault-etal-2019-findings, federmann-etal-2022-ntrex} is an MT benchmark derived from WMT19 news data translated by professional human translators. It contains 1997 parallel sentences and is recommended for the evaluation of from-English translation directions. We use the first 1000 sentence pairs for evaluation, and the last 997 sentence pairs as the selection pool.
    \item \textbf{TICO-19}~\citep{anastasopoulos-etal-2020-tico} is an MT benchmark comprising texts on the COVID-19 pandemic covering 35 languages. Its validation and test sets consist of 971 (used as a selection pool) and 2100 samples respectively.
\end{itemize}
We focus on translating from English. We report the results obtained on NTREX~128 in Table~\ref{tab:ntrex_zs_hrl} and those obtained on TICO-19 in Table~\ref{tab:tico_zs_lrl}. We reach the same conclusions on NTREX~128 than with FLORES 200. On TICO-19 we consider different languages, mostly from Asia. Despite some of them being low-resource languages (Khmer, Marathi, Nepali), thinking offers little to no advantage.

\begin{table*}[ht]
\vskip 0.15in
\small
\begin{center}
\resizebox{\textwidth}{!}{
\begin{tabular}{lrrrrrrrrrrrrrr}
\toprule
\multirow{2}{*}{Models}  & \multicolumn{2}{c}{Czech} & & \multicolumn{2}{c}{Finnish} & & \multicolumn{2}{c}{French} & & \multicolumn{2}{c}{German} & & \multicolumn{2}{c}{Japanese}\\
\cmidrule{2-3} \cmidrule{5-6} \cmidrule{8-9} \cmidrule{11-12} \cmidrule{14-15}
{} & {BLEU} & {MetricX} &  & {BLEU} & {MetricX} & & {BLEU} & {MetricX} & & {BLEU} & {MetricX} & & {BLEU} & {MetricX}\\
\midrule
\textsc{Qwen3-0.6B}           & 4.62 & 22.82 & & 3.03 & 23.26 & & \bf 16.80 & 10.38 & & \bf 13.93 & 10.38 & & \bf 5.66 & 10.06  \\
\textit{\quad w/o Thinking}   & \bf 4.92 & \bf 15.63 & & \bf 3.33 & \bf 21.09 & & 16.72 & \bf 9.88 & & \bf 13.89 & \bf 9.18 & & 4.18 & \bf 8.78 \\
\midrule
\textsc{Qwen3-1.7B}           & 13.64 & 16.44 & & \bf 6.93 & 19.37 & & \bf 26.14 & \bf 5.86 & & \bf 24.33 & \bf 4.97 & & \bf 12.93 & \bf 6.94 \\
\textit{\quad w/o Thinking}   & \bf 14.11 & \bf 16.13 & & \bf 6.95 & \bf 19.19 & & 25.61 & 6.13 & & 23.43 & \bf 4.98 & & 12.41 & 7.03 \\
\midrule
\textsc{Qwen3-4B}             & \bf 22.23 & \bf 9.92 & & \bf 12.70 & \bf 14.04 & & \bf 31.07 & \bf 4.09 & & \bf 30.41 & \bf 2.97 & & \bf 18.11 & \bf 5.64  \\
\textit{\quad w/o Thinking}   & 22.06 & 10.16 & & 11.83 & 14.56 & & 30.82 & 4.43 & & 30.21 & 3.16 & & 16.77 & 5.91 \\
\midrule
\textsc{Qwen3-8B}             & \bf 28.40 & \bf 7.19 & & \bf 17.02 & \bf 10.28 & & 33.10 & \bf 3.51 & & 34.19 & \bf 1.98 & & 19.55 & \bf 5.09 \\
\textit{\quad w/o Thinking}   & 27.54 & 7.92 & & 16.63 & 10.82 & & \bf 33.44 & \bf 3.51 & & \bf 34.27 & 2.20 & & \bf 19.99 & 5.15  \\
\midrule
\textsc{Qwen3-14B}            & \bf 31.04 & \bf 5.68 & & \bf 20.22 & \bf 7.84 & & 34.61 & \bf 3.07 & & \bf 36.85 & \bf 1.68 & & \bf 22.02 & \bf 4.65  \\
\textit{\quad w/o Thinking}   & 29.92 & 6.18 & & 19.86 & 8.25 & & \bf 34.80 & 3.17 & & \bf 36.82 & \bf 1.69 & & 21.88 & \bf 4.61  \\
\midrule
\textsc{Qwen3-32B}            & \bf 32.63 & \bf 5.73 & & 21.32 & 7.47 & & \bf 34.76 & 2.95 & & \bf 37.09 & 1.53 & & \bf 21.19 & 4.97  \\
\textit{\quad w/o Thinking}   & 30.48 & \bf 5.73 & & \bf 21.71 & \bf 7.38 & & 34.55 & \bf 2.84 & & \bf 37.13 & \bf 1.46 & & 20.59 & \bf 4.51  \\
\end{tabular}
}
\resizebox{\textwidth}{!}{
\begin{tabular}{lrrrrrrrrrrrrrr}
\toprule
\multirow{2}{*}{Models}  & \multicolumn{2}{c}{Kazakh} & & \multicolumn{2}{c}{Lithuanian} & & \multicolumn{2}{c}{Portuguese} & & \multicolumn{2}{c}{Spanish} & & \multicolumn{2}{c}{Turkish}\\
\cmidrule{2-3} \cmidrule{5-6} \cmidrule{8-9} \cmidrule{11-12} \cmidrule{14-15}
{} & {BLEU} & {MetricX} &  & {BLEU} & {MetricX} & & {BLEU} & {MetricX} & & {BLEU} & {MetricX} & & {BLEU} & {MetricX}\\
\midrule
\textsc{Qwen3-0.6B}           & 0.26 & 23.60 & & 0.90 & 24.26 & & \bf 19.57 & 10.00 & & \bf 22.01 & 8.77 & & 6.76 & 21.87 \\
\textit{\quad w/o Thinking}   & 0.66 & 22.05 & & 1.46 & 23.90 & & 18.37 & \bf 9.28 & & 20.12 & \bf 8.21 & & \bf 7.04 & \bf 20.30 \\
\midrule
\textsc{Qwen3-1.7B}           & 0.52 & 23.71 & & 4.62 & 21.42 & & \bf 29.23 & \bf 5.49 & & \bf 30.94 & \bf 4.83 & & \bf 12.58 & 16.33 \\
\textit{\quad w/o Thinking}   & 1.13 & 23.68 & & 5.06 & 21.47 & & 28.92 & 5.60 & & 30.81 & 4.96 & & 12.73 & \bf 15.25 \\
\midrule
\textsc{Qwen3-4B}             & 5.95 & 17.29 & & 10.59 & \bf 15.76 & & \bf 34.60 & \bf 3.76 & & \bf 36.44 & \bf 3.32 & & \bf 19.26 & \bf 11.29 \\
\textit{\quad w/o Thinking}   & 5.54 & 17.81 & & \bf 10.86 & 15.83 & & 33.83 & 4.23 & & 35.91 & 3.62 & & 19.05 & \bf 11.31 \\
\midrule
\textsc{Qwen3-8B}             & \bf 9.70 & \bf 13.39 & & 14.80 & 11.97 & & 36.71 & \bf 3.10 & & \bf 38.91 & \bf 2.78 & & 23.07 & \bf 8.67  \\
\textit{\quad w/o Thinking}   & 8.93 & 14.07 & & \bf 14.96 & \bf 11.92 & & \bf 36.89 & 3.35 & & 38.85 & 2.95 & & \bf 23.29 & 8.84  \\
\midrule
\textsc{Qwen3-14B}            & \bf 13.18 & \bf 10.57 & & 18.89 & 9.58 & & \bf 38.49 & \bf 2.83 & & \bf 40.24 & \bf 2.53 & & \bf 26.35 & \bf 7.38 \\
\textit{\quad w/o Thinking}   & 12.65 & 11.04 & & \bf 19.42 & \bf 9.16 & & 38.39 & 2.91 & & \bf 40.18 & 2.59 & & \bf 26.31 & \bf 7.36 \\
\midrule
\textsc{Qwen3-32B}            & \bf 13.53 & 10.93 & & 19.93 & 9.01 & & \bf 39.21 & \bf 2.74 & & 40.70 & 2.41 & & \bf 27.62 & \bf 6.93 \\
\textit{\quad w/o Thinking}   & 13.05 & \bf 10.40 & & \bf 20.77 & \bf 8.62 & & \bf 39.23 & \bf 2.76 & & \bf 41.30 & \bf 2.35 & & 26.68 & 6.97 \\
\bottomrule
\end{tabular}
}
\end{center}
\caption{BLEU and MetricX scores for 10 English $\rightarrow$ X directions from NTREX~128\iffalse~\citep{federmann-etal-2022-ntrex, barrault-etal-2019-findings}\fi. Best results are highlighted in bold.}
\label{tab:ntrex_zs_hrl}
\end{table*}
\begin{table*}[ht]
\vskip 0.15in
\small
\begin{center}
\resizebox{\textwidth}{!}{
\begin{tabular}{lrrrrrrrrrrrrrr}
\toprule
\multirow{2}{*}{Models}  & \multicolumn{2}{c}{Bengali} & & \multicolumn{2}{c}{Farsi} & & \multicolumn{2}{c}{Hindi} & & \multicolumn{2}{c}{Indonesian} & & \multicolumn{2}{c}{Khmer}\\
\cmidrule{2-3} \cmidrule{5-6} \cmidrule{8-9} \cmidrule{11-12} \cmidrule{14-15}
{} & {BLEU} & {MetricX} &  & {BLEU} & {MetricX} & & {BLEU} & {MetricX} & & {BLEU} & {MetricX} & & {BLEU} & {MetricX}\\
\midrule
\textsc{Qwen3-0.6B}           & 0.07 & 22.31 & & 1.86 & 20.03 & & 0.32 & 21.64 & & \bf 22.66 & 7.43 & & 0.63 & 22.87 \\
\textit{\quad w/o Thinking}   & 0.32 & 21.61 & & 3.17 & 3.46 & & 0.58 & 21.79 & & 21.18 & \bf 6.51 & & 1.02 & 23.80 \\
\midrule
\textsc{Qwen3-1.7B}           & 2.85 & 15.49 & & 10.07 & \bf 13.50 & & 6.32 & 14.70 & & 37.04 & \bf 3.47 & & 1.04 & 22.10  \\
\textit{\quad w/o Thinking}   & \bf 4.69 & \bf 13.13 & & \bf 10.46 & 13.56 & & \bf 9.03 & \bf 13.15 & & \bf 38.33 & \bf 3.45 & & 1.54 & 22.21  \\
\midrule
\textsc{Qwen3-4B}             & \bf 14.50 & 6.31 & & \bf 19.11 & \bf 7.62 & & \bf 24.40 & \bf 6.23 & & 45.14 & 2.82 & & \bf 9.20 & \bf 15.05  \\
\textit{\quad w/o Thinking}   & 13.90 & \bf 6.16 & & 18.59 & 7.72 & & 23.40 & 6.34 & & \bf 46.60 & \bf 2.52 & & 8.27 & 15.54  \\
\midrule
\textsc{Qwen3-8B}             & \bf 18.97 & \bf 4.13 & & \bf 25.13 & 4.70 & & \bf 30.22 & \bf 4.82 & & 48.83 & \bf 1.89 & & \bf 16.66 & \bf 10.65 \\
\textit{\quad w/o Thinking}   & 17.98 & \bf 4.15 & & 24.91 & \bf 4.58 & & 29.95 & 4.90 & & \bf 50.19 & 1.94 & & 16.20 & \bf 10.63  \\
\midrule
\textsc{Qwen3-14B}            & \bf 23.68 & \bf 3.10 & & 29.01 & 3.55 & & \bf 35.81 & \bf 4.12 & & 51.21 & \bf 1.66 & & 22.09 & \bf 8.21  \\
\textit{\quad w/o Thinking}   & 22.87 & 3.28 & & \bf 29.33 & \bf 3.38 & & 35.65 & 4.26 & & \bf 52.07 & 1.71 & & \bf 22.30 & 8.41   \\
\midrule
\textsc{Qwen3-32B}            & 17.94 & 7.76 & & 27.39 & 4.41 & & 31.15 & 6.83 & & 51.95 & \bf 1.62 & & 16.52 & 11.42  \\
\textit{\quad w/o Thinking}   & \bf 24.40 & \bf 2.89 & & \bf 29.52 & \bf 3.16 & & \bf 37.37 & \bf 3.98 & & \bf 52.47 & \bf 1.59 & & \bf 21.22 & \bf 7.13   \\
\end{tabular}
}
\resizebox{\textwidth}{!}{
\begin{tabular}{lrrrrrrrrrrrrrr}
\toprule
\multirow{2}{*}{Models}  & \multicolumn{2}{c}{Marathi} & & \multicolumn{2}{c}{Malay} & & \multicolumn{2}{c}{Nepali} & & \multicolumn{2}{c}{Tagalog} & & \multicolumn{2}{c}{Urdu}\\
\cmidrule{2-3} \cmidrule{5-6} \cmidrule{8-9} \cmidrule{11-12} \cmidrule{14-15}
{} & {BLEU} & {MetricX} &  & {BLEU} & {MetricX} & & {BLEU} & {MetricX} & & {BLEU} & {MetricX} & & {BLEU} & {MetricX}\\
\midrule
\textsc{Qwen3-0.6B}           & 0.03 & 23.58 & & 13.57 & 9.11 & & 0.24 & 22.75 & & 5.66 & 21.11 & & 0.41 & 22.99 \\
\textit{\quad w/o Thinking}   & 0.06 & 24.04 & & 10.49 & 4.98 & & 0.61 & 21.58 & & \bf 7.72 & \bf 17.76 & & 0.85 & 22.18  \\
\midrule
\textsc{Qwen3-1.7B}           & 1.27 & 19.50 & & 22.81 & 4.71 & & 2.45 & 17.30 & & 10.18 & \bf 18.67 & & 2.74 & 18.42 \\
\textit{\quad w/o Thinking}   & \bf 2.49 & \bf 18.36 & & \bf 23.17 & \bf 4.54 & & \bf 9.83 & \bf 10.04 & & \bf 10.65 & \bf 18.64 & & 3.08 & 18.26 \\
\midrule
\textsc{Qwen3-4B}             & \bf 7.78 & \bf 9.80 & & 31.93 & \bf 3.99 & & \bf 9.83 & \bf 10.04 & & 21.62 & 11.54 & & \bf 11.37 & \bf 10.82 \\
\textit{\quad w/o Thinking}   & \bf 7.75 & 10.61 & & \bf 32.87 & 4.07 & & 8.13 & 11.42 & & \bf 24.30 & \bf 10.52 & & 10.55 & 10.97 \\
\midrule
\textsc{Qwen3-8B}             & \bf 12.01 & \bf 6.92 & & 37.30 & \bf 2.95 & & \bf 15.25 & \bf 6.93 & & 27.44 & 7.60 & & 16.35 & 7.06  \\
\textit{\quad w/o Thinking}   & 11.93 & 7.03 & & \bf 39.25 & 3.12 & & 14.42 & 7.53 & & \bf 30.00 & \bf 7.18 & & \bf 16.56 & \bf 6.88  \\
\midrule
\textsc{Qwen3-14B}            & 15.05 & \bf 5.35 & & 41.91 & \bf 2.63 & & \bf 18.23 & \bf 5.89 & & 32.38 & \bf 5.35 & & \bf 20.97 & \bf 5.06 \\
\textit{\quad w/o Thinking}   & \bf 15.11 & 5.49 & & \bf 42.91 & \bf 2.67 & & 17.93 & 6.26 & & \bf 34.29 & \bf 5.33 & & \bf 21.04 & 5.24 \\
\midrule
\textsc{Qwen3-32B}            & 13.26 & 8.61 & & 43.30 & \bf 2.57 & & 17.40 & 8.31 & & 31.44 & 6.04 & & 20.21 & 6.36 \\
\textit{\quad w/o Thinking}   & \bf 16.63 & \bf 4.98 & & \bf 44.12 & 2.65 & & \bf 20.83 & \bf 5.45 & & \bf 36.12 & \bf 4.80 & & \bf 21.87 & \bf 4.87  \\
\bottomrule
\end{tabular}
}
\end{center}
\caption{BLEU and MetricX scores for 10 English $\rightarrow$ X directions from TICO-19\iffalse~\citep{anastasopoulos-etal-2020-tico}\fi. Best results are highlighted in bold.}
\label{tab:tico_zs_lrl}
\end{table*}

\subsection{Does distilled Chain-of-Thought as intermediate tokens improve performance?} \label{appendix:cot_distillation_lt}

\begin{figure}[ht]
\begin{center}
    \includegraphics[width=\linewidth]{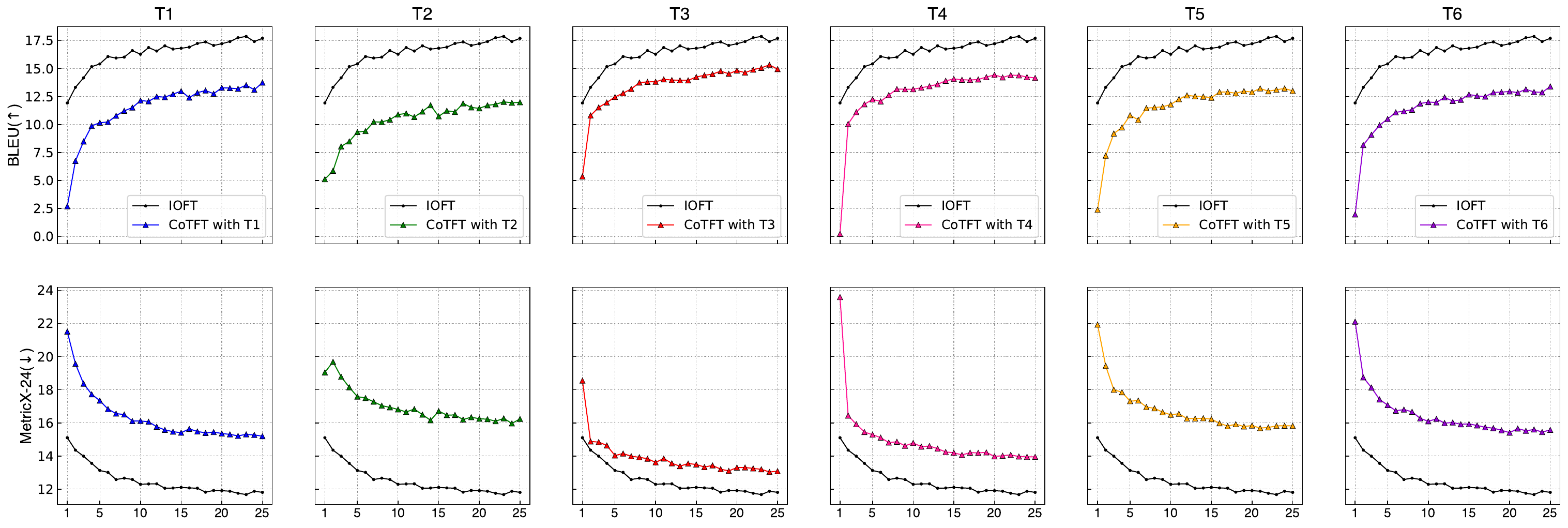}
    \caption{Comparison between \textsc{CoTFT} and \textsc{IOFT} with six different CoT templates.}
    \label{fig:coft_lt}
\end{center}
\end{figure}

Following Section~\ref{subsection:cot_distillation}, we compare \textsc{CoTFT} with \textsc{IOFT} across all six templates using \texttt{gemma-3-1b-pt} as the student and \texttt{gemma-3-27b-it} as the intermediate teacher. We focus on translating from English to Lithuanian. As shown in Figure~\ref{fig:coft_lt}, \textsc{CoTFT} consistently lags behind \textsc{IOFT}. The gap can be as large as 5 BLEU and 4 MetricX. Despite T3 being the best template, it is still largely behind \textsc{IOFT} in terms of performance.

\subsection{What happens when we use traces from MT prompting strategies as intermediate tokens?} \label{appendix:prompting_lt}

\begin{figure}[ht]
\begin{center}
    \includegraphics[width=\linewidth]{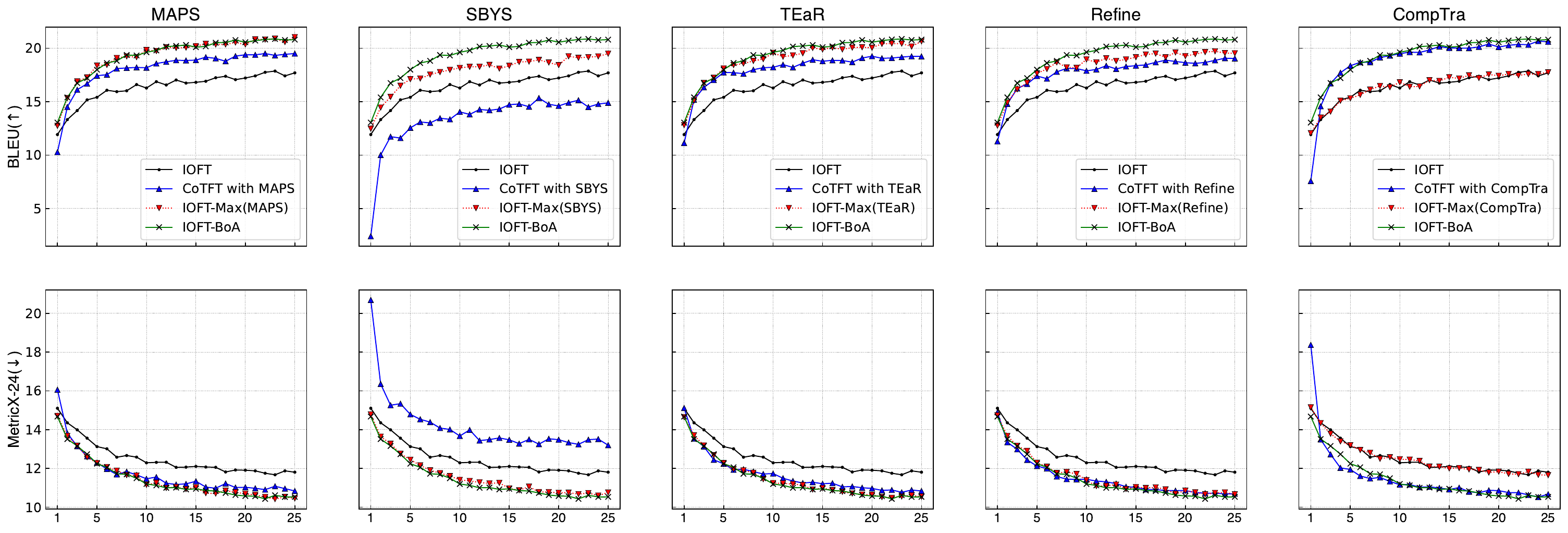}
    \caption{Comparison between \textsc{IOFT} and \textsc{CoTFT} with five different prompting strategies.}
    \label{fig:prompting_lt}
\end{center}
\end{figure}

In Figure~\ref{fig:prompting_lt}, we observe that \textsc{CoTFT} with reasoning traces based on alternative prompting strategies outperforms \textsc{IOFT}. SBYS is an exception, for which \textsc{CoTFT} is behind \textsc{IOFT}. Across prompting strategies, \textsc{IOFT-Max} outperforms \textsc{IOFT} and \textsc{CoTFT} with the only exception of CompTra. This is exactly what happened with our experiments with \texttt{Llama-4-Scout-17B-16E-Instruct} and \texttt{gemma-3-4b-pt} in Xhosa.

\begin{figure}[ht]
\begin{center}
    \includegraphics[width=\linewidth]{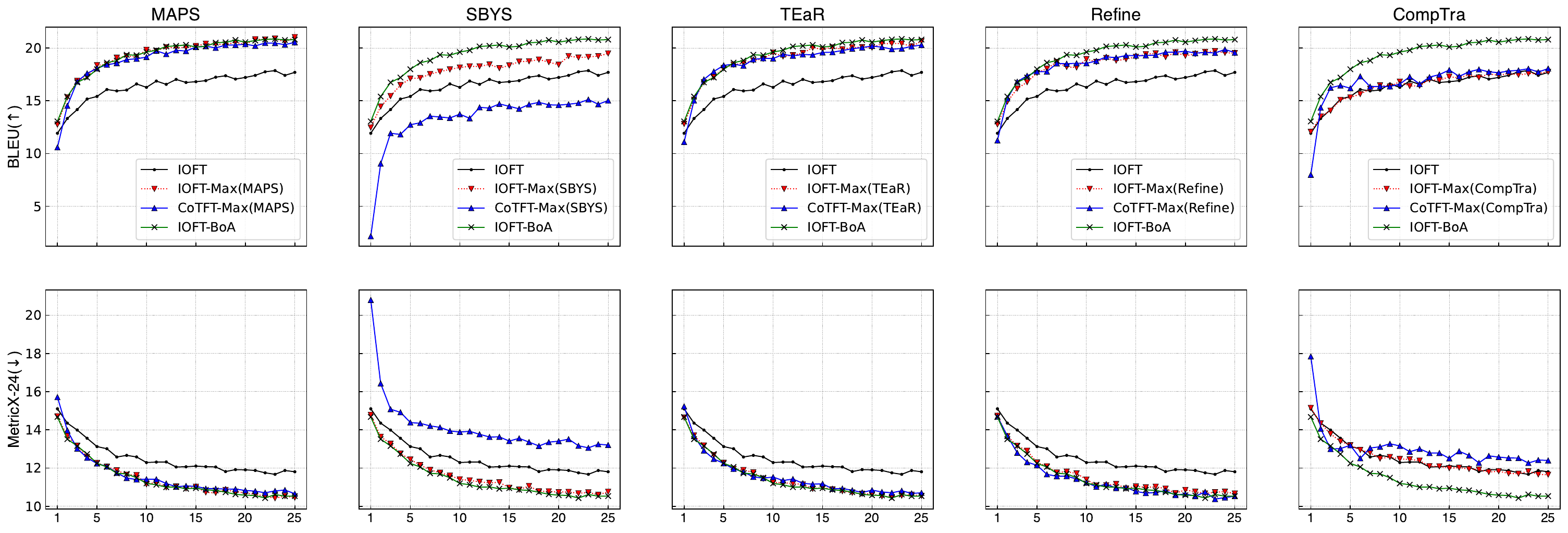}
    \caption{Comparison between \textsc{IOFT} and \textsc{CoTFT} with five different prompting strategies.}
    \label{fig:prompting_lt_v2}
\end{center}
\end{figure}

Finally, as shown in Figure~\ref{fig:prompting_lt_v2}, \textsc{CoTFT-Max} fails to improve over \textsc{IOFT-Max}, confirming our previous conclusions with Xhosa. In this case, CompTra is not an exception. For all the strategies, \textsc{CoTFT-Max} and \textsc{IOFT-Max} are very close in performance, with \textsc{IOFT-BoA} topping them all. This again suggests that reasoning traces do not help, even when they are based on MT prompting strategies whose drafting attempts do not outperform the ground truth in terms of quality. Having target translations of high-quality (\textsc{IOFT-BoA}) has the highest impact, outperforming the standard \textsc{IOFT} by 3 BLEU and 1.3 MetricX with the same number of parallel pairs and the same training recipe.

\subsection{Down the rabbit hole of sentence decomposition} \label{appendix:sentence_decomposition_lt}

\begin{figure}[ht]
\begin{center}
    \includegraphics[width=\linewidth]{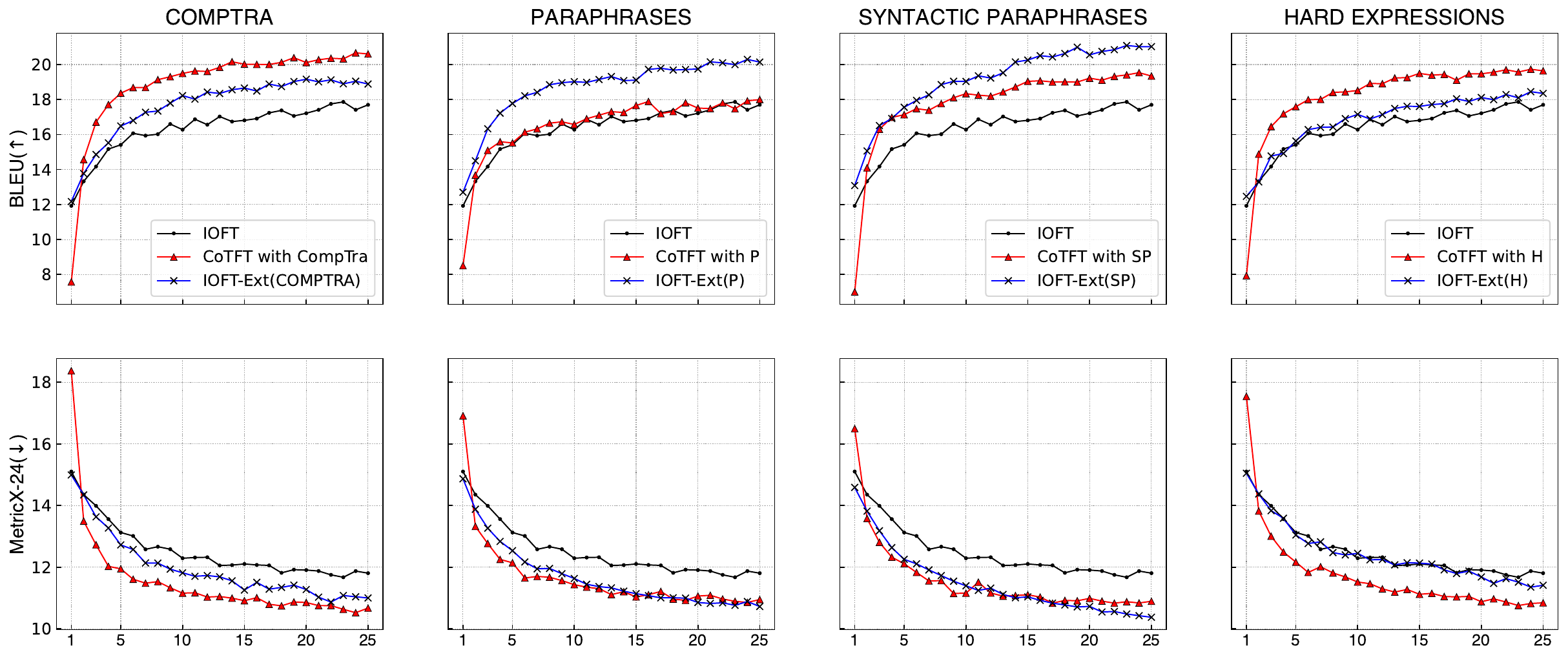}
    \caption{Comparison between \textsc{IOFT} and \textsc{CoTFT} with four different sentence decomposition strategies.}
    \label{fig:paraphrase_v2}
\end{center}
\end{figure}

Following Section~\ref{subsection:sentence_decomposition}, we evaluate multiple sentence decomposition approaches and compare \textsc{CoTFT} against standard \textsc{IOFT} and \textsc{IOFT-Ext}. As shown in Figure~\ref{fig:paraphrase_v2}, \textsc{CoTFT} consistently outperforms \textsc{IOFT} across all decomposition strategies. Again, SP and CompTra works better than P and H. Using the generated pairs as additional training samples (i.e. \textsc{IOFT-Ext}) is particularly helpful with P and SP because they correspond to fully-fledged sentences as explained earlier. \textsc{CoTFT} with H and \textsc{CoTFT} with CompTra outperform the corresponding \textsc{IOFT-Ext} suggesting that short phrases and their translations are relevant intermediate information for \textsc{CoTFT}. \textsc{CoTFT} with CompTra works just as well as \textsc{IOFT-Ext(P)} which is impressive as the latter required multiplying the size of $\mathcal{D}$ by six.

\subsection{CoT distillation: What happens when we change the teacher?} \label{appendix:thinking_deepseek}

\begin{figure}[ht]
\begin{center}
    \includegraphics[width=\linewidth]{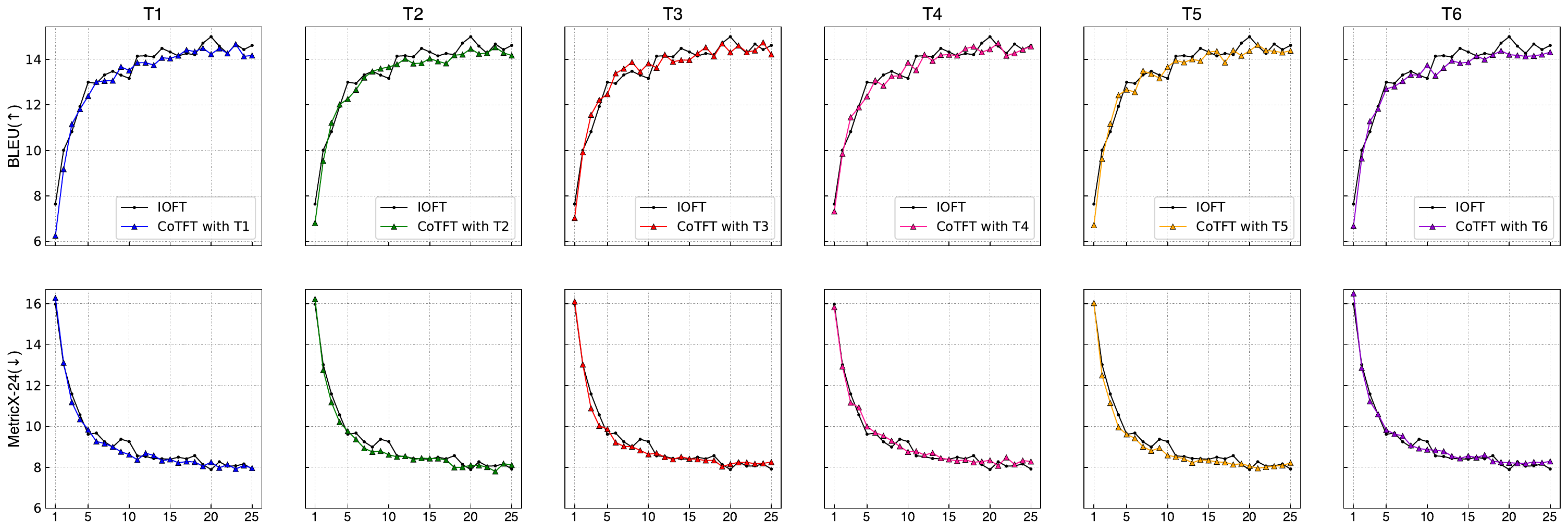}
    \caption{Comparison between \textsc{IOFT} and \textsc{CoTFT} with six different CoT templates.}
    \label{fig:coft_ds}
\end{center}
\end{figure}

In this section, we run the same CoT distillation experiment as in Section~\ref{subsection:cot_distillation} but we use \texttt{DeepSeek-R1-Distill-Llama-70B} instead as the teacher. As seen in Figure~\ref{fig:coft_ds}, \textsc{CoTFT} behaves similarly to \textsc{IOFT} across templates. The performance of \textsc{CoTFT} is better than what we observed with \texttt{Llama-4-Scout-17B-16E-Instruct} despite \textsc{LLaMA} being better at translating into Xhosa. We attribute this to the ``thinking'' abilities of \textsc{DeepSeek-R1} which, despite not being good at generating Xhosa, can generate a better explanation compared to \textsc{LLaMA} as to why a hypothesis is an accurate translation of a source. However, in both cases, \textsc{CoTFT} does not improve over \textsc{IOFT}, which is also faster to train in comparison.

\subsection{MT traces generated by prompting strategies as intermediate tokens: What happens when we change the teacher?} \label{appendix:impact_of_intermediate_teacher}
We run additional information on Xhosa and change the teacher from \texttt{Llama-4-Scout-17B-16E-Instruct} to \texttt{gemma-3-27b-it}. \textsc{Gemma}'s performance zero-shot MT performance (on FLORES 200, BLEU=12.82, MetricX=7.62) is worse than \textsc{LLaMA}'s (BLEU=16.90, MetricX=6.53) and we want to investigate how this impacts our findings. First of all, \textsc{CoTFT} outperforms \textsc{IOFT} across prompting strategies with the exception of SBYS. \textsc{IOFT-Max} outperforms \textsc{IOFT} and we observe the same behaviour between \textsc{IOFT-Max(CompTra)} and \textsc{IOFT} as we did with \textsc{LLaMA}.
Despite \textsc{CoTFT} with SBYS underperforming \textsc{IOFT}, \textsc{IOFT-Max(SBYS)} outperforms \textsc{IOFT}, meaning that translation attempts embedded in SBYS-inspired CoT are helpful, but they are drowned out by other useless tokens, which impacts how well \textsc{CoTFT} performs. Ultimately, \textsc{IOFT-BoA} works best; although \textsc{IOFT-Max(TEaR)} achieves higher BLEU scores it lags behind in terms of MetricX. It even outperforms the teacher \texttt{gemma-3-27b-it} despite being six times smaller.

\begin{figure}[ht]
\begin{center}
    \includegraphics[width=\linewidth]{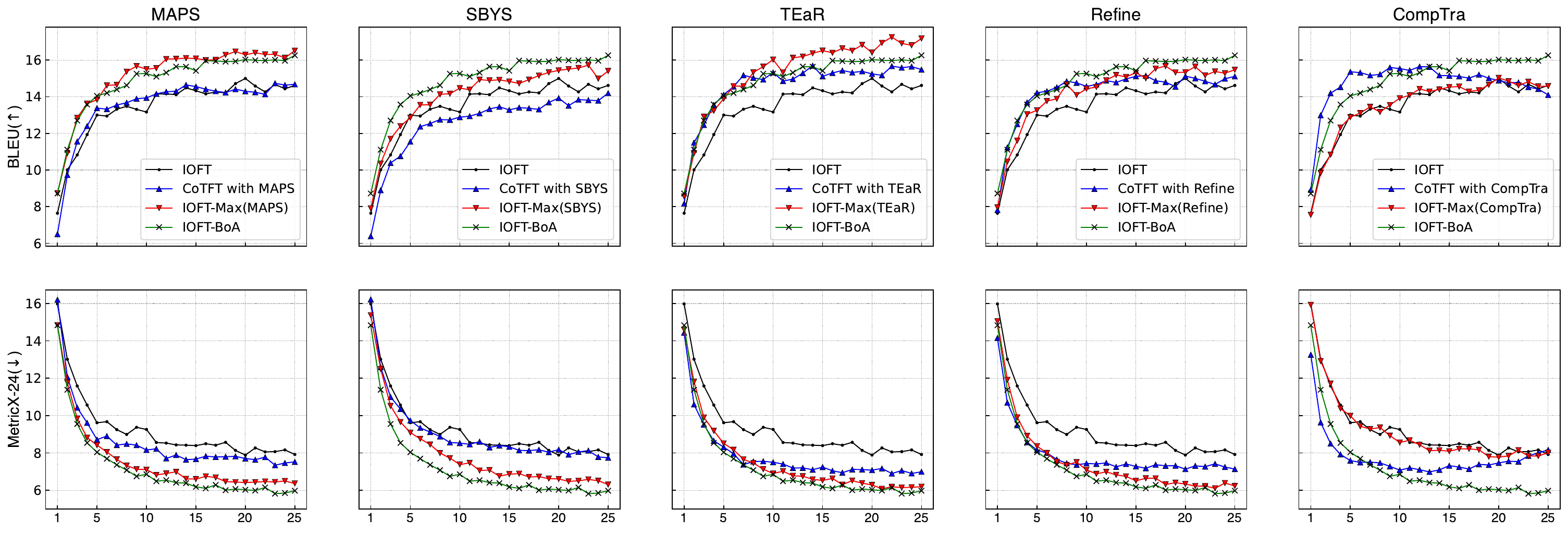}
    \caption{Comparison between \textsc{IOFT} and \textsc{CoTFT} with five different prompting strategies.}
    \label{fig:prompting_v4}
\end{center}
\end{figure}

\begin{figure}[ht]
\begin{center}
    \includegraphics[width=\linewidth]{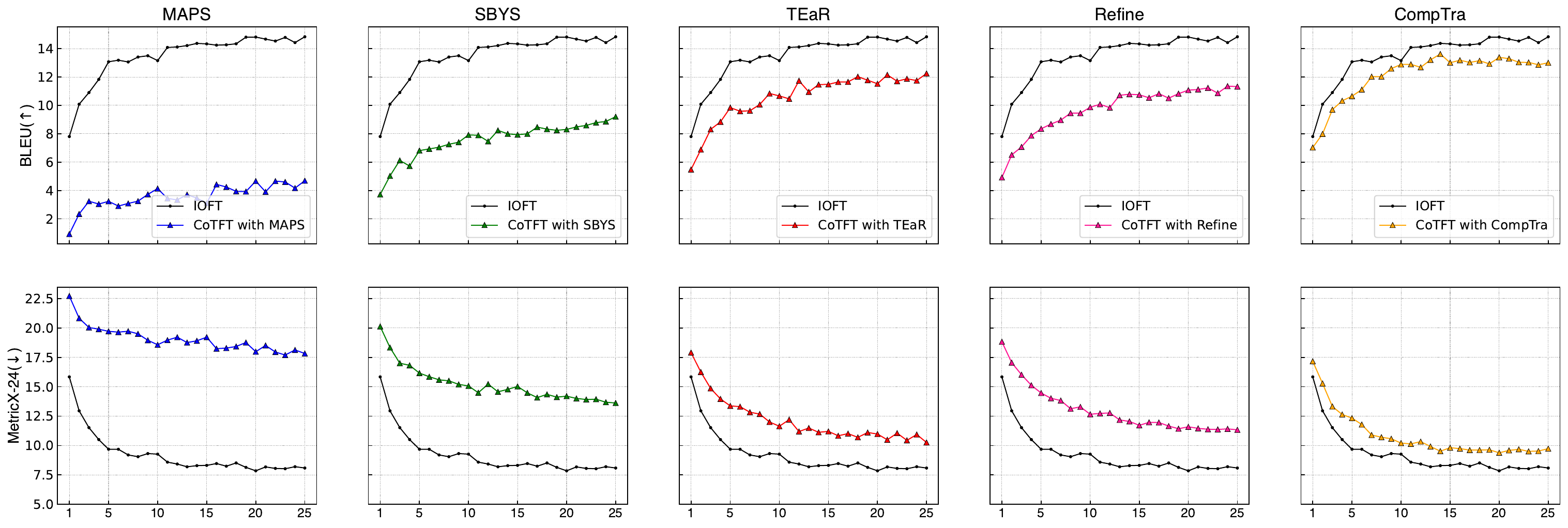}
    \caption{Comparison between \textsc{IOFT} and \textsc{CoTFT} with five different prompting strategies.}
    \label{fig:prompting_v5}
\end{center}
\end{figure}

When we use \texttt{gemma-3-4b-it} as the teacher (\texttt{gemma-3-4b-pt} being the student), the traces obtained using prompting strategies do not help \textsc{CoTFT} to outperform \textsc{IOFT}. In Figure~\ref{fig:prompting_v5}, we observe a degradation of performance that confirms our intuition suggesting that these traces are helpful only if they contain translation attempts that are better than the ground truth.

\subsection{Reinforcement Learning After IOFT and CoFT} \label{appendix:rl}
\begin{figure}[ht]
\begin{center}
    \includegraphics[height=8cm]{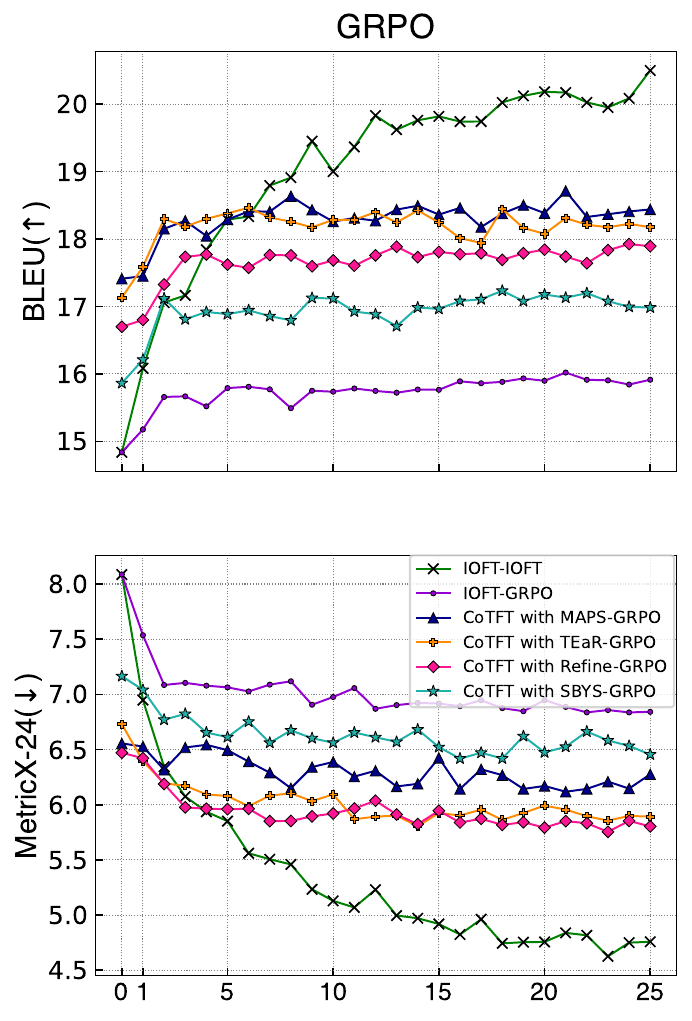}
    \caption{Comparison between \textsc{IOFT} and \textsc{CoTFT} with GRPO.}
    \label{fig:coft_rl_ext}
\end{center}
\end{figure}

Building on the experiments presented in Section~\ref{section:rl}, we apply GRPO to the final checkpoints (checkpoint-5000) under three additional configurations: \textsc{CoTFT} with MAPS, SBYS, TEaR, and Self-Refine. The results, shown in Figure~\ref{fig:coft_rl_ext}, indicate that GRPO results in consistent improvements of approximately +1 BLEU and -0.7 MetricX points across all setups, mirroring the trends observed with CompTra. Notably, GRPO maintains the relative performance ordering between \textsc{IOFT} and \textsc{CoTFT} prior to fine-tuning. However, \textsc{CoTFT} models do not gain more from GRPO than \textsc{IOFT} models, and in practice, performing \textsc{IOFT} alone (rather than GRPO) can achieve comparable or greater gains at a lower computational cost.

\end{document}